\documentclass[10pt,journal,compsoc,pdftex]{IEEEtran}

\usepackage{hyperref}
\hypersetup{
    colorlinks=true,
    linkcolor=blue,
    filecolor=magenta,      
    urlcolor=cyan,
    pdfpagemode=FullScreen,
    }
\usepackage[export]{adjustbox}

\ifCLASSOPTIONcompsoc
  \usepackage[nocompress]{cite}
\else
  \usepackage{cite}
\fi

\ifCLASSINFOpdf
   \usepackage[pdftex]{graphicx}
\else
   \usepackage[dvips]{graphicx}
\fi

\usepackage{amsmath}
\usepackage{amsfonts}
\usepackage{algorithmic}

\usepackage{array}

\usepackage{subfigure}
\usepackage{tikz}

\usepackage{cleveref}
\crefrangelabelformat{figure}%
      {#3#1#4--#5\crefstripprefix{#1}{#2}#6}
\crefrangelabelformat{subfigure}%
      {#3#1#4#5-(\crefstripprefix{#1}{#2}#6}
\Crefname{figure}{Fig.}{Figs.}

\usepackage{multirow}
\usepackage{floatflt}
\usepackage{tabularx}
\usepackage{booktabs}
\usepackage[export]{adjustbox} 
\usepackage{bbm}
\usepackage{dsfont} 

\usepackage{float}
\usepackage{xspace}
\usepackage{siunitx} 
\usepackage{etoolbox}
\usepackage{balance}
\usepackage{colortbl}

\usepackage{xcolor}
\usepackage{soul}

\renewrobustcmd{\bfseries}{\fontseries{b}\selectfont}

\ifCLASSOPTIONcaptionsoff
 \usepackage[nomarkers]{endfloat}
\let\MYoriglatexcaption\caption
\renewcommand{\caption}[2][\relax]{\MYoriglatexcaption[#2]{#2}}
\fi

\newcommand{\argmax}[1]{\underset{#1}{\operatorname{arg}\,\operatorname{max}}\;}
\newcommand{\cifar}{\textsc{Cifar}\xspace}  
\newcommand{\webface}{\textsc{CASIA}-WebFace\xspace}

\newcommand{\normnummecc}{$AC$\xspace} 
\newcommand{\avgmetr}{$AM$\xspace}
\newcommand{\normnummeccformula}{AC} 
\newcommand{\avgmetrformula}{AM} 

\makeatletter
\renewcommand*\env@matrix[1][\arraystretch]{%
  \edef\arraystretch{#1}%
  \hskip -\arraycolsep
  \let\@ifnextchar\new@ifnextchar
  \array{*\c@MaxMatrixCols c}}
\makeatother

\begin{document}
\title{CoReS: Compatible Representations \\ via Stationarity}

\author{Niccolò~Biondi, 
        Federico~Pernici, 
        Matteo~Bruni, 
        and~Alberto~Del~Bimbo,~\IEEEmembership{Senior~Member,~IEEE}
\thanks{
The authors are with Media Integration and Communication Center (MICC), Dipartimento di Ingegneria dell'Informazione, Università degli Studi di Firenze, 50139, Firenze, Italy. E-mail: name.surname@unifi.it
}
}


\IEEEtitleabstractindextext{%
\begin{abstract}
Compatible features enable the direct comparison of old and new learned features allowing to use them interchangeably over time. In visual search systems, this eliminates the need  to extract new features from the gallery-set when the representation model is upgraded with novel data. This has a big value in real applications as re-indexing the gallery-set can be computationally expensive when the gallery-set is large, or even infeasible due to privacy or other concerns of the application. In this paper, we propose CoReS, a new training procedure to learn representations that are \textit{compatible}  with those previously learned, grounding on the stationarity of the features as provided by fixed classifiers based on polytopes. With this solution, classes are maximally  separated in the representation space and maintain their spatial configuration stationary as new classes are added, so that there is no need to learn any mappings between representations nor to impose pairwise training with the previously learned model.  
We demonstrate that our training procedure largely outperforms the current state of the art and is particularly effective in the case of multiple upgrades of the training-set, which is the typical case in real applications.
\end{abstract}

\begin{IEEEkeywords}
Deep Convolutional Neural Network, Representation Learning, Compatible Learning, Fixed Classifiers. 
\end{IEEEkeywords}
}

\maketitle

\IEEEdisplaynontitleabstractindextext

\IEEEpeerreviewmaketitle

\section{Introduction} \label{sec:intro}

\IEEEPARstart{N}{atural}
intelligent systems learn from visual experience and seamlessly exploit such learned knowledge to identify similar entities. Modern artificial intelligence systems, on their turn, typically require distinct phases to perform such visual search. 
An internal representation is first learned from a set of images (the \emph{training-set}) using Deep Convolutional Neural Network models (DCNNs) \cite{chopra2005learning,bengio2013representation,sharif2014cnn,YosinskiNIPS2014} and then used to index a large corpus of images (the \textit{gallery-set}). Finally, visual search is obtained by identifying the closest images in the gallery-set to an input \emph{query-set} by comparing their representations. 
Successful applications of learning feature representations are: face-recognition \cite{schroff2015facenet,liu2016large,taigman2014deepface,deepID_NIPS2014,DBLP:conf/cvpr/DengGXZ19}, person re-identification \cite{yi2014deep,li2014deepreid,zheng2017person,chen2019abd}, image retrieval \cite{babenko2014neural,gordo2016deep,tolias2021particular}, and car re-identification \cite{khan2019survey} among the others.

\begin{figure}[t]
    \centering
    \includegraphics[width=0.99\linewidth]{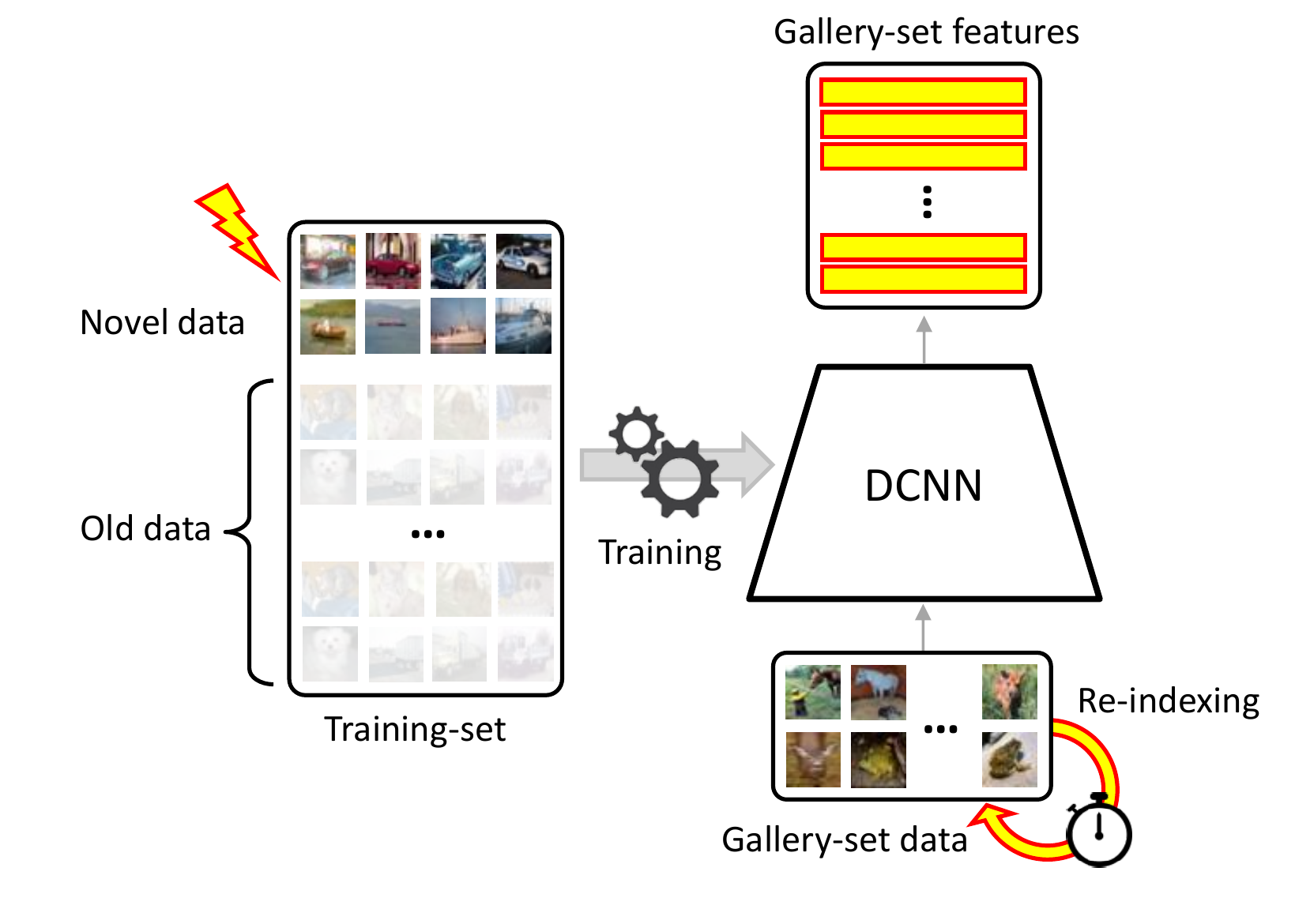}
    
    \caption{
    Upgrading the DCNN representation model with novel data, typically requires the gallery-set to be re-indexed. Learning compatible representations allows to compare the newly learned representation of an input query-set with the old representation of the gallery-set, thus eliminating its computationally intensive re-indexing.}
    \label{fig:model_updates_comparison}
\end{figure}

In the case in which novel data for the training-set and/or more recent or powerful network architectures become available, the representation model may require to be \textit{upgraded} to improve its search capabilities. In this case, not only the query-set but also all the images in the gallery-set should be re-processed by the upgraded model to generate new features and replace the old ones to benefit from such upgrading. The re-processing of the gallery-set is referred to as \emph{re-indexing}  (Fig.~\ref{fig:model_updates_comparison}).

For visual search systems with a large gallery-set, such as in surveillance systems, social networks or in autonomous robotics,
re-indexing is clearly computationally expensive \cite{DBLP:conf/cvpr/ShenXXS20} or has critical deployment, especially when the working system requires multiple upgrades or there are real-time constraints.
Re-indexing all the images in the gallery-set can be also infeasible when, due to privacy or ethical concerns, the original gallery images cannot be permanently stored \cite{van2020ethical} and the only viable solution is to continue using the feature vectors previously computed. 
In all these cases, it should be possible to directly compare the upgraded features of the query with the previously learned features of the gallery, i.e., the new representation should be \emph{compatible} with the previously learned representation.

Learning compatible representation has recently received increasing attention and novel methods have been proposed in \cite{DBLP:conf/cvpr/ShenXXS20}, \cite{duggal2021compatibility}, \cite{Chen_2019_CVPR}, \cite{hu2019towards}, \cite{DBLP:conf/bmvc/WangCYCL20}, \cite{Meng_2021_ICCV}. Differently from these works, in this paper we address \textit{compatibility} leveraging the \textit{stationarity} of the learned internal representation.  Stationarity allows to maintain the same distribution of the features over time so that it is possible to compare the features of the upgraded representation with those previously learned. In particular, we enforce stationarity by leveraging the properties of a family of classifiers whose parameters are not subject to learning, namely \emph{fixed classifiers} based on regular polytopes \cite{pernici2020icpr}\cite{Pernici2019CVPRW}\cite{perniciTNNLS2021}, that allow to reserve regions of the representation space to future classes while classes already learned remain in the same spatial configuration. 

The main contributions of our research are the following:
\begin{enumerate}
    \item We identify stationarity as a key property for compatibility and propose a novel training procedure for learning compatible feature representations via stationarity, without the need of learning any mappings between representations nor to impose pairwise training with the previously learned model. We called our method: Compatible Representations via Stationarity (CoReS).
    \item We introduce new criteria for comparing and evaluating compatible representations in the case of sequential multi-model upgrading.
    \item We demonstrate through extensive evaluation on large scale verification, re-identification and retrieval benchmarks that CoReS improves the current state-of-the-art in learning compatible features for both single and sequential multi-model upgrading.
\end{enumerate}

In the following, in Sec.~\ref{sec:rel_works}, we discuss the main literature on compatible representation learning and highlight the distinguishing features of our solution.
In  Sec.~\ref{sec:Problem Statement and Evaluation}, we present in detail the problem of learning compatible representations and define new criteria and metrics for compatibility evaluation in sequential multi-model upgrading.
In Sec.~\ref{sec:proposed-stationarity}, we present our solution for learning compatible representations by exploiting feature stationarity. 
In Sec.~\ref{sec:experiments}, we evaluate our solution against state-of-the-art methods on different benchmark datasets and network architectures and demonstrate its superior performance in learning compatible representations. Finally, in Sec.~\ref{sec:ablation_studies}, we perform an extensive ablation study.

\section{Related Works} \label{sec:rel_works}
\noindent
\textbf{Compatible Representation Learning}. 
The term \textit{backward compatibility} was first introduced in \cite{bansal2019updates} for the classification task. 
They noted that although  machine learning models can increase on average their performance with the availability of more training data, 
upgrading the model could result into incorrect classification of data correctly classified with the previous model. As a consequence, the trust in machine learning systems is severely harmed. 
Compatibility in classification has been further investigated in
\cite{srivastava2020an},
\cite{yan2021positive},
\cite{trauble2021backward},
\cite{gygli2021towards}. 

However, learning compatible representations is substantially different from learning compatible classifier models, although both follow the same general principle. As a distinct hallmark, learning compatible representations directly imposes constraints in the semantic distance of the feature representation. 
In \cite{Chen_2019_CVPR}, \cite{hu2019towards},  \cite{DBLP:conf/bmvc/WangCYCL20}, \cite{Meng_2021_ICCV} the problem of feature compatibility was addressed by learning a \textit{mapping} between two representation models so that the new and old feature vectors can be directly compared. 
The mapping in \cite{Chen_2019_CVPR} was learned through a three-step procedure: adversarial learning for reconstruction, feature extraction and regression to jointly optimize the whole model. In \cite{hu2019towards}, the mapping was learned through an autoencoder by minimizing the distance between the two representation spaces and the reconstruction error. 
In \cite{DBLP:conf/bmvc/WangCYCL20}, the mapping was learned from a residual bottleneck transformation module trained by three different losses: classification loss, similarity loss between feature spaces, and KL-divergence loss between the prototypes of the classifiers. 
In \cite{Meng_2021_ICCV}, the estimated mapping aligns the
class prototypes between the models. To further encourage compatibility, the method also reduces intra-class variations for the new model.
All these methods do not completely avoid the cost of re-indexing of the gallery-set as, at each upgrade, the old feature vectors must be re-processed with the learned mapping. Therefore, they are not suited for sequential multi-model learning and large gallery-sets. 
Differently from these works, we avoid learning specific space-to-space mappings for each previous upgraded representation model and completely avoid the cost of re-indexing also in the case of multiple upgrades. 

By avoiding to learn space-to-space mappings, our work has some affinity with the  Backward Compatible Training (BCT) \cite{DBLP:conf/cvpr/ShenXXS20} method for compatible learning that represents the current state-of-the-art. 
BCT grounds on pairwise compatibility learning to obtain compatible features. It takes advantage of an influence loss that biases the new representation in a way that it can be used by both the new and the old classifier. During learning with novel data, the old classifier is \textit{fixed} and the prototypes of the new classifier align with the prototypes of the old classifier. 
In the case of multi-model upgrading, such pairwise cooperation supports compatibility only indirectly (i.e., through transitive compatibility). In fact, for a two-model upgrading,  i.e., when the model $\phi_1$ is upgraded to $\phi_2$ and $\phi_2$ to $\phi_3$, $\phi_3$ is  compatible with $\phi_1$ thanks to the compatibility of $\phi_3$ with $\phi_2$ and of $\phi_2$ with $\phi_1$.
BCT has been extended in \cite{duggal2021compatibility}, where small and large representation models are taken into account for the query and the gallery-set, respectively. 
Differently from BCT our method is not based on pairwise compatibility learning and does not use previous classifiers which might be incorrectly learned. Instead, we learn a representation that is directly compatible to \textit{all} the previous representations by following a training strategy that only leverages feature stationarity.

A different compatible representation learning scenario, referred to as asymmetric metric learning, was addressed in \cite{budnik2021asymmetric}. Large network architectures are used for the gallery-set and smaller architectures for queries without considering new data for the training-set. 

Compatibility was implicitly studied also in \cite{li2015convergent,NEURIPS2018_5fc34ed3} in which representation similarity between two networks with identical architecture but trained from different initializations was evaluated.

\noindent
\textbf{Neural Collapse}.
Our method is based on the concept of learning stationary and maximally separated features using the $d$-Simplex \textit{fixed} classifier introduced in \cite{Pernici2019CVPRW} and \cite{perniciTNNLS2021}. In this classifier, weights are not trainable and are determined from the coordinates of the vertices of the \mbox{$d$-Simplex regular polytope}. 
The goal of learning stationary and maximally separated features has similarities to the Neural Collapse phenomenon described in \cite{papyan2020prevalence}. This phenomenon, which can be further explored in \cite{kothapalli2022neural}, shows that the final \textit{learnable} classifier and the learned feature representation tend to orient towards a $d$-Simplex structure. Along this line of research, the works \cite{graf2021dissecting} and \cite{zhu2021geometric} investigated fixing the final classifier in the scenario of balanced datasets. In contrast, in \cite{fang2021exploring} it is shown that training on imbalanced datasets does not lead to Neural Collapse. However, \cite{yang2022we} shows that Neural Collapse can occur in imbalanced and long-tail scenarios as long as the classifier is \textit{fixed} to a \mbox{$d$-Simplex}, and this evidence has been recently supported in \cite{kasarla2022maximum} also for out-of-distribution detection. 
Our experimental results on large long-tailed datasets with distribution shifts, such as \cite{weyand2020google} and \cite{ypsilantis2021met}, demonstrate that learning stationary and maximally separated features is effective even in scenarios with increased complexity due to the extra challenge of finding compatible representations. 

\noindent
\textbf{Class-incremental Learning}. 
Class-incremental Learning (CiL) is the process of sequentially increasing the number of classes to learn over time \cite{Hsu18_EvalCL,van2019three,masana2020class}. Although apparently similar to sequential learning of compatible representations, the main focus of CiL is to reduce catastrophic forgetting \cite{mccloskey1989catastrophic}. 
CiL differs from compatible representations learning in two important aspects: (1) in CiL the new model is typically initialized with the old model, while in compatible representation learning this is optional (2) in CiL the new model has not access to the whole data during the model upgrade, while in compatible representation learning all training data is available at each upgrade step.
According to this, in compatible representation learning, the learned representation is not affected by catastrophic forgetting. 

\section{Compatibility Evaluation}
\label{sec:Problem Statement and Evaluation}

We indicate with \mbox{${I}_\mathcal{G}=\{\mathbf{x}_i\}_{i=1}^N$} and $F_\mathcal{G}=\{\mathbf{f}_i\}_{i=1}^N$ respectively the set of images and their features in the gallery-set $\mathcal{G}$. The gallery-set $\mathcal{G}$ might be grouped into a number of classes or identities $L$ according to a set of labels $\mathcal{Y}=\{ y_i \}_{i=1}^L$. We assume that the features $F_\mathcal{G}$ are extracted using the representation model $\phi_{\rm old}:{\mathbb R}^D \rightarrow {\mathbb R}^d$ that transforms each image $\mathbf{x} \in {\mathbb R}^D$ into a feature vector $\mathbf{f} \in {\mathbb R}^d$, where $d$ and $D$ are the dimensions of the feature and the image space, respectively. Analogously, we will refer to  \mbox{${I}_\mathcal{Q}$} and $F_\mathcal{Q}$ respectively as the set of images and their features in the query-set $\mathcal{Q}$. The model $\phi_{\rm old}$ is trained on a training-set $\mathcal{T}_{\rm old}$ and used to perform search tasks using a distance ${\rm dist}:{\mathbb R}^d \times {\mathbb R}^d \rightarrow \mathbb{R}_+$ to identify the closest images to the query images ${I}_\mathcal{Q}$.
As novel images $\mathcal{X}$ become available, a new training-set $\mathcal{T}_{\rm new}=\mathcal{T}_{\rm old} \cup \mathcal{X}$ is created and exploited to learn a new model \mbox{$\phi_{\rm new}:{\mathbb R}^D \rightarrow {\mathbb R}^d$} that improves (i.e., upgrades) the $\phi_{\rm old}$ model.
Our goal is to design a training procedure to learn a compatible model $\phi_{\rm new}$ so that any query image transformed with it can be used to perform search tasks against the gallery-set directly without re-indexing, i.e., without computing\mbox{ $F_\mathcal{G}=\{ \mathbf{f} \in \mathbb{R}^d \, | \,  \mathbf{f} = \phi_{\rm new}(\mathbf{x}) \, \forall \, \mathbf{x} \in I_\mathcal{G}\}$}. 

\subsection{Compatibility Criterion}\label{sec:eval-compatibility}
In \cite{DBLP:conf/cvpr/ShenXXS20}, a general criterion to evaluate compatibility was defined. According to this, a new and compatible representation model must be at least as good as its previous version in clustering images from the same class 
and separating those from different classes.
A new representation model $\phi_{\rm new}$ is therefore compatible with an old representation model $\phi_{\rm old}$ if:
\begin{equation}\label{eq:compatible_pairwise}
\resizebox{0.9\hsize}{!}{$%
\begin{aligned}
 {\rm dist} \big(\phi_{\rm new}(\mathbf{x}_u), \phi_{\rm old}(\mathbf{x}_v) \big) &\leq {\rm dist} \big(\phi_{\rm old}(\mathbf{x}_u), \phi_{\rm old}(\mathbf{x}_v) \big) \\
 & \forall \, (u,v) \in \big\{(u,v) \, | \, y_u = y_v \big\} \\  \mathrm{and} &\\
  {\rm dist} \big(\phi_{\rm new}(\mathbf{x}_u), \phi_{\rm old}(\mathbf{x}_v) \big) &\geq {\rm dist} \big(\phi_{\rm old}(\mathbf{x}_u), \phi_{\rm old}(\mathbf{x}_v) \big) \\
 & \forall \, (u,v) \in \big\{(u,v) \, | \, y_u \neq y_v \big\}, \\
\end{aligned}
$}
\end{equation}
where $\mathbf{x}_u$ and $\mathbf{x}_v$ are two input samples and ${\rm dist}(\cdot,\cdot)$ is a distance in feature space. 
Since it constrains all pairs of samples, Eq.~\ref{eq:compatible_pairwise} 
is relaxed to the following \textit{Empirical Compatibility Criterion}:
\begin{equation} \label{eq:compatible_set}
{M} \big(\phi_{\rm new}^{\cal Q}, \phi_{\rm old}^{\cal G} \big) > {M} \big(\phi_{\rm old}^{\cal Q}, \phi_{\rm old}^{\cal G} \big),
\end{equation}
where $M$ is a metric used to evaluate the performance based on ${\rm dist}(\cdot,\cdot)$. 
The notation ${M} \big(\phi_{\rm new}^{\cal Q}, \phi_{\rm old}^{\cal G} \big)$ underlines that the upgraded model $\phi_{\rm new}$ is used to extract feature vectors $F_\mathcal{Q}$ from query images $I_\mathcal{Q}$, while the old model $\phi_{\rm old}$ is used to extract features $F_\mathcal{G}$ from gallery images $I_\mathcal{G}$. This performance value is referred to as \textit{cross-test}. 
Correspondingly, ${M} \big(\phi_{\rm old}^{\cal Q}, \phi_{\rm old}^{\cal G} \big)$ evaluates the case in which both query and gallery features are extracted with $\phi_{\rm old}$ and is referred to as \textit{self-test}. The underlying intuition of Eq. \ref{eq:compatible_set} is that model $\phi_{\rm new}$ is compatible with $\phi_{\rm old}$ when the cross-test is greater than the self-test, i.e., by using the upgraded representation for the query-set and the old representation for the gallery-set the system improves its performance with respect to the previous condition.

To evaluate the relative improvement gained by a new learned compatible representation, the following \emph{Update Gain} has been defined: 
\begin{equation} \label{eq:update_gain}
{\Gamma} \big(\phi_{\rm new}^{\cal Q}, \phi_{\rm old}^{\cal G} \big) = 
\frac{{M} \big(\phi_{\rm new}^{\cal Q}, \phi_{\rm old}^{\cal G} \big) 
- {M} \big(\phi_{\rm old}^{\cal Q}, \phi_{\rm old}^{\cal G} \big) }
{ {M} \big(\widetilde{\phi}_{\rm new}^{\cal Q}, \widetilde{\phi}_{\rm new}^{\cal G}\big) - {M} \big(\phi_{\rm old}^{\cal Q}, \phi_{\rm old}^{\cal G} \big)},
\end{equation}
where 
$
{M} \big(\widetilde{\phi}_{\rm new}^{\cal Q}, \widetilde{\phi}_{\rm new}^{\cal G}\big)
$ 
stands for the best accuracy level we can achieve by re-indexing the gallery-set with the new representation \cite{DBLP:conf/cvpr/ShenXXS20} and can be considered as the upper bound of the best achievable performance.

\begin{figure}[t]
    \centering
     \includegraphics[width=0.99\linewidth]{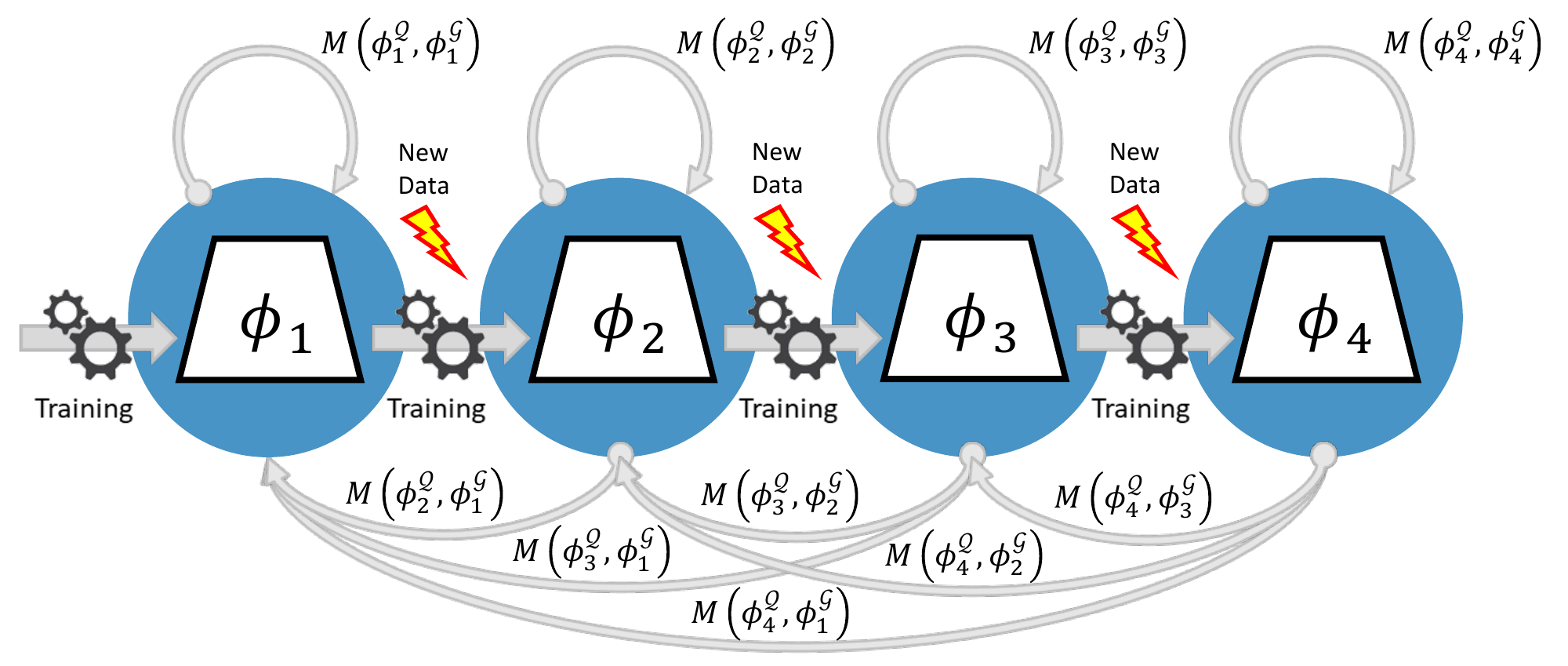}
    \caption{Multi-model Empirical Compatibility Criterion (Eq.~\ref{eq:multistepecc}):  representation models $\phi_i$ with  $i=1,2, \dots,T$ are sequentially trained. Gray arrows represent self and cross-tests (example with $T=4$). 
    }
    \label{fig:multi_model_comp}
\end{figure}

\subsection{Multi-step Compatibility Criterion}
In real world applications, multi-step upgrading is often required, i.e., different representation models must be sequentially learned through time, in multiple upgrade steps. 
At each step $t$, the training-set is upgraded as:
\begin{equation}\label{eq:growing_training_set}
    \mathcal{T}_{t} = \mathcal{T}_{t-1} \cup \mathcal{X}_t
\end{equation}
being $\mathcal{X}_t$ the new data and $\mathcal{T}_{t-1}$ the training-set at step $t-1$.
In the multi-step upgrading case, we define the following \textit{Multi-model Empirical Compatibility Criterion} as follows:
\begin{align}
\label{eq:multistepecc}
\begin{split}
M \big(  \phi_{t'}^\mathcal{Q}, \phi_{t}^\mathcal{G} \big) & >  M \big( \phi_{t}^\mathcal{Q}, \phi_{t}^\mathcal{G} \big)  \quad \; \forall \, t' > t \\
\text{with } t'  \in \big\{2,3, \dots, & T \big\}  \text{ and } t \in \big\{1,2, \dots, T-1 \big\} 
\end{split}
\end{align}
where $\phi_{t'}$ and $\phi_{t}$ are two different models such that $\phi_{t}$ is upgraded before $\phi_{t'}$, $T$ is the number of upgrade steps and $M$  the  metric  used to evaluate the performance.
Model $\phi_{t'}$ is compatible with $\phi_{t}$ when their cross-test is greater than the self-test of $\phi_{t}$ for each pair of upgrade steps.
Fig.~\ref{fig:multi_model_comp} illustrates the Multi-model Empirical Compatibility Criterion, where $\{\phi_1, \phi_2, \ldots, \phi_T\}$ are the representation models, black arrows indicate the model upgrades and gray arrows represent self and cross-tests.

In order to assess multi-model compatibility of Eq.~\ref{eq:multistepecc} for a sequence of $T$ upgrade steps, we define the following square triangular \textit{Compatibility Matrix} $C$:
\begin{align}
 & C = 
 \begin{pmatrix}[1.8]
  M \big( \phi_1^\mathcal{Q}, \phi_1^\mathcal{G} \big) &  &  \\
  M \big( \phi_2^\mathcal{Q}, \phi_1^\mathcal{G} \big) &  M \big( \phi_2^\mathcal{Q}, \phi_2^\mathcal{G} \big)  &   &   \\
  \vdots  & \vdots  & \ddots &   \\
  M \big( \phi_T^\mathcal{Q}, \phi_1^\mathcal{G} \big) & M \big( \phi_T^\mathcal{Q}, \phi_2^\mathcal{G} \big) & \cdots & M \big( \phi_T^\mathcal{Q}, \phi_T^\mathcal{G} \big)
 \end{pmatrix} 
\label{eq:compatibility_matrix}   
\end{align}
where each entry $C_{ij}$ is the performance value according to metric $M$, taking model $\phi_i$ for the query-set $\mathcal{Q}$ and model $\phi_j$ for the gallery-set $\mathcal{G}$. Entries on the main diagonal, $i=j$, represent the self-tests, while the entries off-diagonal, $i > j$, represent the cross-tests. While showing compatibility performance across multiple upgrade steps, matrix $C$ can be used to provide a scalar metric to quantify the global multi-model compatibility in a sequence of upgrade steps. In particular, we define the \textit{Average Multi-model Compatibility} (\normnummecc) as the number of times that Eq.~\ref{eq:multistepecc} is verified
with respect to all its possible occurrences, independently of the number of the learning steps:
\begin{equation}
\small
\normnummeccformula = \frac{2}{T(T-1)}
\sum_{i=2}^T \sum_{j=1}^{i-1}
\mathds{1}{ \Big ( C_{ij} > C_{jj} } \Big ),
\label{eq:AC}
\end{equation}
where $\mathds{1}(\cdot)$ denotes the indicator function.

Finally, we define the \textit{Average Multi-model Accuracy} (\avgmetr) as the average of the entries of the Compatibility Matrix:
\begin{equation}\label{eq:avg_M}
\avgmetrformula = \frac{2}{T(T+1)}
\sum_{i=1}^T \sum_{j=1}^{i}
C_{ij}
\end{equation}
to provide an aggregate value of the accuracy metric $M$ under compatible training.

\section{Learning Compatible Representations }
\label{sec:proposed-stationarity}
It is well known that for different initializations a neural network learns the same subspaces but with different basis vectors \cite{li2015convergent, NEURIPS2018_5fc34ed3}. 
Therefore, training the network from scratch with different randomly initialized weights \textit{does not} provides similar representations in terms of subspace geometry. This result excludes compatibility between two independently trained representation models.

The alternative of learning with incremental fine-tuning (i.e., weights are initialized from the previously learned model) appears to be a more favorable training procedure to compatibility. However, and perhaps counterintuitively, this does not help to keep the same subspace representation geometry regardless of the changes made.

\begin{figure}
    \hspace{-13pt}
    \subfigure[]{ \label{fig:feat_space_change_pre}
        \adjincludegraphics[width=0.48\linewidth,trim={0 0 {.5\width} 0},clip]{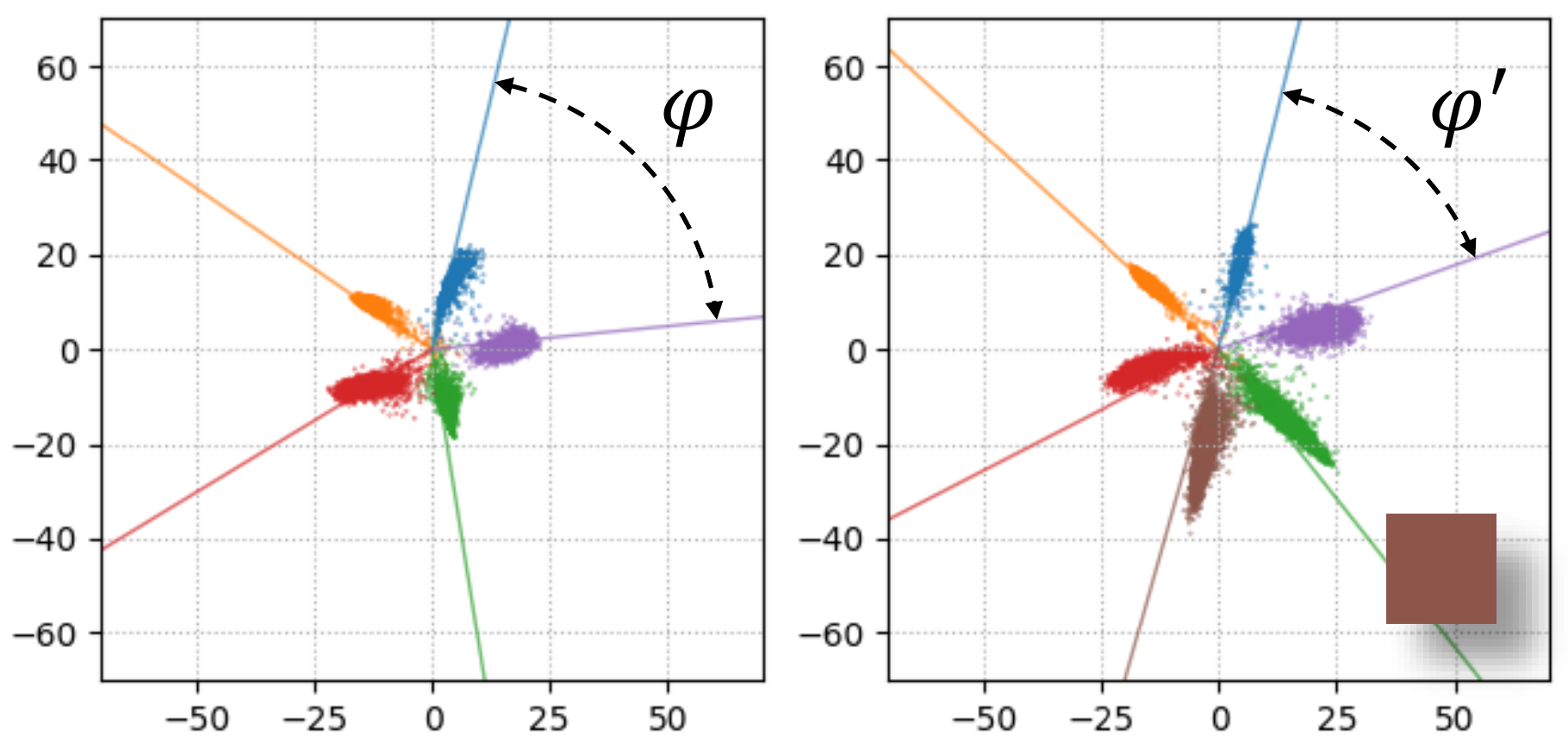}
    }
    \subfigure[]{ \label{fig:feat_space_change_post}
        \adjincludegraphics[width=0.48\linewidth,trim={{.5\width} 0 0 0},clip]{imgs/figure2_adding_a_class.pdf}
    }
    \caption{
    Learning with incremental fine-tuning with MNIST dataset for 2D representation. Colored cloud points represent features from the test-set and gray lines represent classifier prototypes. (a) Initial configuration (5 classes); (b) Training by fine-tuning  (adding the brown-class). The addition of the new class modifies the spatial configuration and angles between features.
    }    \label{fig:feat_space_change}
\end{figure}

We provide direct evidence of this aspect of feature learning in Fig.~\ref{fig:feat_space_change} with a 
toy problem. We trained the LeNet++ architecture \cite{wen2016discriminative} on a subset of the MNIST dataset setting the output size of the last hidden layer to two (so resulting in a two-dimensional representation space).
The classifier weights were unit normalized and biases were set to zero to encourage learning cosine distance between features (\cite{DBLP:conf/icml/LiuWYY16,Liu2017CVPR}) and the cross-entropy loss was used. The model was initially trained with five classes (red, orange, blue, purple, and green clouds in Fig.~\ref{fig:feat_space_change_pre}); then a new class (brown cloud in Fig.~\ref{fig:feat_space_change_post}) was included in the training-set and the new model was trained by fine-tuning the old model on the new training-set of six classes.
As the new class is included in the training-set and the representation is fine-tuned, the features of the old classes change their spatial configuration and the mutual angles between classifier prototypes change as well.
This is due to the fact that linear classifiers maximize inter-class distance to better discriminate between classes
\cite{wen2016discriminative}.
As a consequence, the cosine distance comparison between old and new features cannot be guaranteed. The same effect holds for any number of classes and feature space dimension. 

To limit such spatial configuration changes and therefore achieve feature compatibility, our approach learns stationary features exploiting the properties of fixed classifiers introduced in \cite{perniciTNNLS2021} that we briefly recall in the next subsection. 

\begin{figure*}[t]
    \centering
    \includegraphics[width=0.9\textwidth]{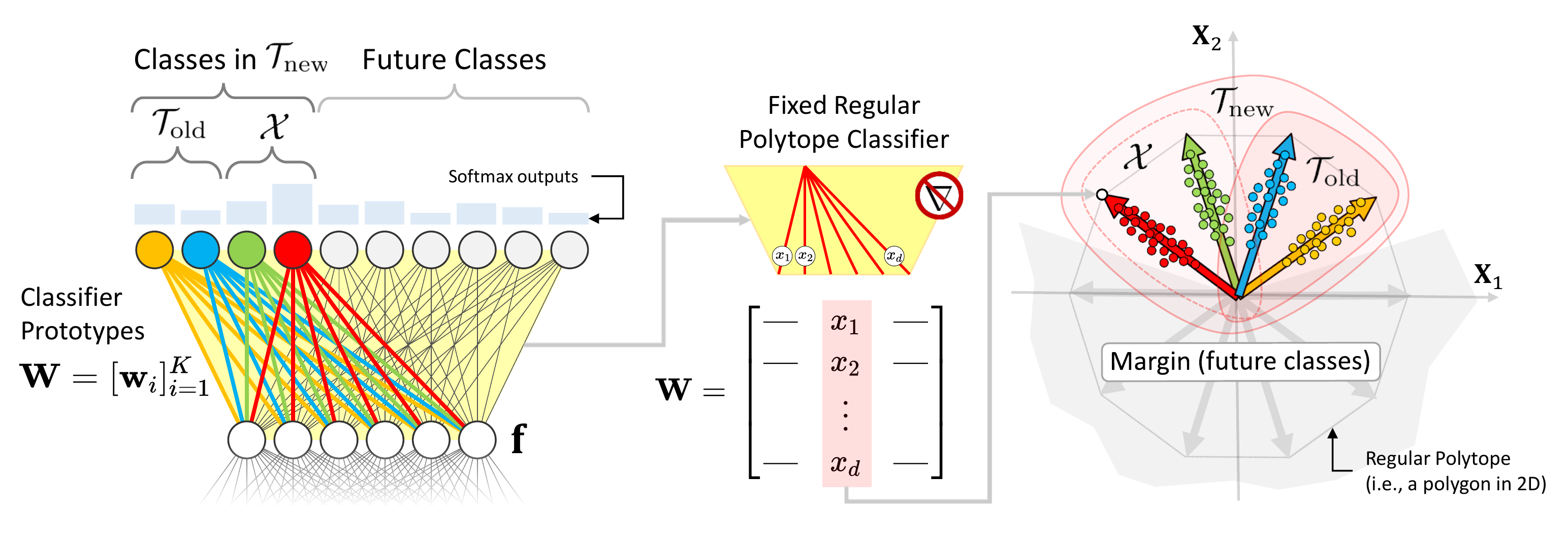}
    \subfigure[]{\hspace{0.35\textwidth}\label{fig:reponet_(a)}}
    \subfigure[]{\hspace{0.2\textwidth}\label{fig:reponet_(b)}}
    \subfigure[]{\hspace{0.25\textwidth}\label{fig:reponet_(c)}}
    \caption{
    Overview of the training procedure of CoReS based on feature stationarity showing: \textit{(a)} fixed classifier with class prototypes $[\mathbf{w}_i]_{i=1}^K$ with evidence of those reserved for $\mathcal{T}_{\rm old}$ and the upgrade classes  $\mathcal{X}$ (both create $\mathcal{T}_{\rm new}$), and those reserved for future classes; \textit{(b)} class prototypes and their parameters $x_1,x_2, \dots x_d$ (parameters are the coordinate vertices of the regular polytope that defines the fixed classifier);
    \textit{(c)} two-dimensional representation of the feature space generated by the fixed classifier. The colored point clouds represent the learned features. Prototypes of the future classes are represented with gray arrows. The gray region is the margin imposed by the future/unseen classes. As new features are learned they are pushed out from the margin.
    }
    \label{fig:reponet}
\end{figure*}

\subsection{Learning Stationary Features with Fixed Classifiers }
In \cite{DBLP:conf/iclr/HardtM17, DBLP:conf/iclr/HofferHS18}, a DCNN model with a fixed classification layer (i.e., not subject to learning) initialized by random weights was shown to be almost equally effective as a trainable classifier with substantial saving of computational and memory requirements. 
In fixed classifiers, the functional complexity of the classifier is fully demanded to the internal layers of the neural network. As the parameters of the classifier prototypes are not trainable, only the feature vector directions align toward the directions of the prototypes.
In \cite{perniciTNNLS2021}, we presented a special class of fixed classifiers where the weights of the classifier are fixed to values taken from the coordinates of the vertices of regular polytopes. Regular polytopes generalize regular polygons in any number of dimensions, and reflect the tendency of splitting the available space into approximately equiangular regions. There are only three possible polytopes in a multi-dimensional feature space with dimensions 5 and higher: the $d$-Simplex, the $d$-Cube and the $d$-Orthoplex.
In these classifiers, the spatial configuration of the learned classes remains stationary as new classes are added and, at the same time, classes are maximally separated in the representation space \cite{pernici2020icpr}.

\subsection{Learning Compatibility via Stationarity}
\label{sec:learning_with_cores}

Compatibility has a close relationship with feature stationarity. In fact, stationarity requires that the representation that is learned in the future is statistically indistinguishable from the  representation learned in the  past, regardless  of  the number of model upgrades performed over time.
We therefore argued that a compatibility training procedure should be defined by directly exploiting stationarity of the representation as provided by fixed classifiers.

According to this, in order to maintain the compatibility across model upgrades, at each upgrade we learn the features of the new classes in reserved regions of the representation space, while the features learned in the previous upgrades will not ``move'' thanks to the stationarity property.
Region reservation is made at the beginning of training by setting a number of classifier outputs greater than the number of the initial classes and keeping not assigned regions for future upgrades. Fig.~\ref{fig:reponet} provides an overview of the CoReS procedure for compatible representation learning. Fig.~\ref{fig:reponet_(a)} shows the initial training-set $\mathcal{T}_{\rm old}$ (two classes for simplicity), the new data $\mathcal{X}$ (two classes for simplicity), the regions left for the future classes, the class prototypes $\mathbf{w}_i$, and the softmax outputs of the fixed classifier. As a result, the training-set $\mathcal{T}_{\rm new}$ that will be used to upgrade the initial representation has four classes. 
Fig.~\ref{fig:reponet_(c)} shows the representation space of a 10-sided 2D regular polytope (a polygon) with the old and new classes, their prototypes and class samples. The regions reserved for the future classes are colored in gray. Fig.~\ref{fig:reponet_(b)} highlights the coordinates of one of the fixed classifier prototypes.

\makeatletter
\newcommand{\vist}{\bBigg@{2.5}}
\newcommand{\vast}{\bBigg@{3.5}}
\newcommand{\Vast}{\bBigg@{5}}
\makeatother

As new data $\mathcal{X}_t$ is available at time $t$ to upgrade the representation, CoReS performs optimization of model $\phi_t$ according to the following loss: 
\begin{eqnarray}
    \mathcal{L}_t=
    -\dfrac{1}{N_b} 
    \sum\limits_{i=1}^{N_b} \log \!  \vast( 
    \dfrac {e^{ {\mathbf{W}}_{y_i}^{\top}{{\phi_t(\mathbf{x}_i)}} }} {\sum\limits_{\scriptscriptstyle j=1}^{\scriptscriptstyle 
    K_t
    } e^{  {\mathbf{W}}_{j}^{\top}{{\phi_t(\mathbf{x}_i)}} } + \sum\limits_{\scriptscriptstyle j = K_t
    +1}^{\scriptscriptstyle K }e^{  {\mathbf{W}}_{j}^{\top}{{\phi_t(\mathbf{x}_i)}} }} \vast)
    \label{softmax_loss_virtual},
\end{eqnarray}
where: $\mathbf{x}_i \in \mathcal{T}_{t-1} \cup \mathcal{X}_t$ is a sample instance; $y_i \in \{1,2,\dots, K_t \}$ its class; $\phi_t(\mathbf{x}_i) \in \mathbb{R}^d$ its feature vector as from model $\phi_t$; $\mathbf{W} \in \mathbb{R}^{d \times K}$ the fixed classifier weight matrix; $K_t$ and $K$, respectively the number of classes in $\mathcal{T}_{t-1} \cup \mathcal{X}_t$ and the number of classifier outputs; and $N_b$ the number of samples in the mini-batch.

For the $K - K_t$ outputs that have not yet been associated to classes at time $t$, the classifier responds with false positives of the training-set $\mathcal{T}_t$. To highlight the role of these outputs, in the denominator of the loss in Eq.~\ref{softmax_loss_virtual}, we distinguished the contributions corresponding to the $K_t$ classes at time $t$ (the first summation) from the $K - K_t$ future classes (the second summation). 
The contribution of the future classes enforces an angular margin penalty in the representation space that tends to push away the features already learned from the prototypes of future classes. 
Such a margin penalty has a key role for learning compatible features as, when upgrading the model with new data, it prevents the new data to affect the representation already learned.
As is evident from Eq.~\ref{softmax_loss_virtual}, CoReS learns feature compatibility without using the previously learned model or classifier.

In our implementation, CoReS uses the $d$-Simplex fixed classifier as it exploits the only polytope that provides class prototypes equidistant from each other \cite{perniciTNNLS2021}.
In this way, we avoid bias in the  assignment of class labels to classifier outputs. 
The $d$-Simplex fixed classifier defined in a feature space of dimension $d$ can accommodate a number of classes equal to its number of vertices, i.e., $K = d+1$.
As a result, the $K$ classifier prototypes are computed as:
\begin{align}
\mathbf{W} =
\vist [e_1,e_2,\dots,e_{K-1}, 
{ 
\footnotesize
\text{$\frac{1-\sqrt{K}}{K-1}$} 
} \sum_{i=1}^{K-1} e_i \vist ]
\label{eq:d-simplex}
\end{align}
where $e_i$ denotes the standard basis in $\mathbb{R}^{d}$, with $i \in \{1,2, \dots, K-1\}$. 
Increasing the dimensionality of the representation has no effect on both the training time and memory consumption, since fixed classifiers do not require back-propagation on the fully connected layers \cite{DBLP:conf/iclr/HofferHS18,perniciTNNLS2021}.

\subsection{CoReS Extension by Model Selection} 
\label{sec:extensions_cores}
At each upgrade, the training-set grows. This causes a shift in the distribution between the training-set and the gallery-set \cite{quinonero2009dataset}. 
To cope with this shift, we extended CoReS with a simple model selection strategy, based on cross validation. 

We assume that: 
(1) drifting occurs gradually such that the density ratio of the marginal distribution before and after the upgrade is close to uniform \cite{sugiyama2007covariate};
(2) there is a set of data  $\mathcal{\widehat{G}}$ available that is an i.i.d. sample of the gallery-set data $\mathcal{G}$ (i.e., a split following the same distribution);
(3) the previously learned model is available. 
Under these assumptions, at each upgrade step, for each epoch, we search the model with the highest self-test among the models that satisfy the compatibility criterion of Eq.~\ref{eq:compatible_set} with respect to the previous model:
\begin{equation}
\label{eq:argmax} 
    \begin{aligned}
\phi_{t'} = \; & \argmax{l} &   M\left( \phi^\mathcal{Q}_{t',l}, \phi^\mathcal{\widehat{G}}_{t',l} \right) \qquad \qquad \qquad \qquad \; & \\
        \; & \text{s. t.} &  M \left( \phi^\mathcal{Q}_{t',l}, \phi^\mathcal{\widehat{G}}_{t} \right) > M\left( \phi^\mathcal{Q}_{t}, \phi^\mathcal{\widehat{G}}_{t} \right) \qquad &
    \end{aligned}
\end{equation}
\noindent
where $t = t' - 1$ and $l$ is the index of the training epoch. 

\begin{figure*}[t]
    \centering
    \begin{minipage}{0.95\textwidth}
        \subfigure[ITM]{\label{fig:3step_upperbound}
            \includegraphics[width=0.140\linewidth]{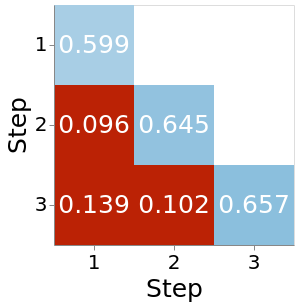}
        }
        \hspace{-2pt}
        \subfigure[$\ell^2$]{\label{fig:3step_lf}
            \includegraphics[width=0.140\linewidth]{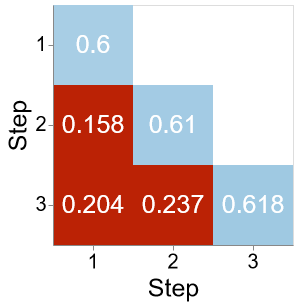}
        }
        \hspace{-2pt}
        \subfigure[IFT]{ \label{fig:3step_ift}
            \includegraphics[width=0.140\linewidth]{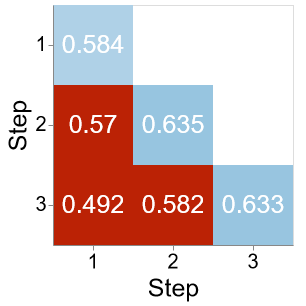}
        }
        \hspace{-2pt}
        \subfigure[LwF~\cite{DBLP:conf/eccv/LiH16}]{ \label{fig:3step_lwf}
            \includegraphics[width=0.140\linewidth]{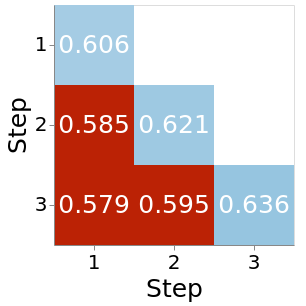}
        }
        \hspace{-2pt}
        \subfigure[BCT~\cite{DBLP:conf/cvpr/ShenXXS20}]{ \label{fig:3step_bct}
            \includegraphics[width=0.140\linewidth]{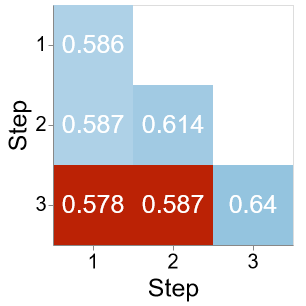}
        }
        \hspace{-2pt}
        \subfigure[CoReS]{ \label{fig:3step_cores}
            \includegraphics[width=0.140\linewidth]{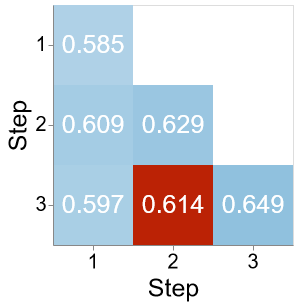}
        }
        \hspace{-2pt}
        \subfigure{
            \subfigcapmargin = 0pt
            \includegraphics[width=0.030\linewidth]{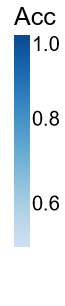}
            \addtocounter{subfigure}{-1}
        }
    \end{minipage}
 
\caption{Compatibility matrices for open set verification on \cifar-100/10 with 2-step multi-model upgrading with different methods compared. Models are sequentially learned on \cifar-100 with 33\%, 66\%, 100\% of data. Diagonal and off-diagonal elements respectively report the self-test and the cross-test accuracy. Models that do not satisfy the compatibility criterion are highlighted in red. }
\label{fig:multi_step_with_baseline}

\end{figure*}

\begin{table} [t]
\small
\centering 
\caption{ 
Compatibility comparative evaluation for open set verification on \cifar100/10 with one upgrade. Models are learned with 50\% and upgraded with 100\% of training data. Subscripts indicate the method used for compatibility learning. Columns indicate accuracy ($M$), whether the Empirical Compatibility Criterion is verified ($ECC$), the Update Gain ($\Gamma$) and the Absolute Gain.} 
\label{tab:baselinecomp} 
\setlength{\tabcolsep}{4.5pt}
\sisetup{detect-weight,mode=text,table-format=1.2,round-precision=2,round-mode=places}
\begin{tabular}{l  S c c S}  
\toprule 
\multirow{2}{*}{\shortstack{\textsc{Comparison Pair}}} 
& {\multirow{2}{*}{\shortstack{$M$}}} 
& \multirow{2}{*}{\shortstack{\textsc{ECC}}}  
& \multirow{2}{*}{\shortstack{$\Gamma$ (\%)}} 
& {\multirow{2}{*}{\shortstack{\textsc{Absolute} \\ \textsc{Gain}}}} \\ 

& \multicolumn{1}{c}{ }  \\ 

\midrule
$(\phi_{\rm old}, \phi_{\rm old})$ {\footnotesize(Lower Bound)} & 0.5973 & \textendash  & \textendash  & \textendash  \\
\midrule
{$(\phi_{{\rm new}-{\rm \textsc{ITM}}}, \phi_{\rm old})$} & {0.12} & {$\times$} & {\textendash}  & {\textendash}  \\
$(\phi_{{\rm new}-\ell^2}, \phi_{\rm old})$ & 0.345 & $\times$ & \textendash  & \textendash  \\
$(\phi_{{\rm new-IFT}}, \phi_{\rm old})$ & 0.5662 & $\times$ & \textendash  & \textendash  \\
$(\phi_{{\rm new-LwF}}, \phi_{\rm old})$ & 0.5885 & $\times$ & \textendash  & \textendash  \\
$(\phi_{{\rm new-BCT}}, \phi_{\rm old})$ & 0.5993 & $\surd$ & 5.9 & 0.2 \\
$(\phi_{{\rm new-CoReS}}, \phi_{\rm old})$ {\footnotesize}& \bfseries 0.6078 & $\surd$ & \textbf{21.3} & \bfseries 1.05 \\
\midrule
$(\widetilde{\phi}_{\rm new}, \widetilde{\phi}_{\rm new})$ {\footnotesize(Upper bound)} & 0.6465 & \textendash  & \textendash  & 0.491 \\
\bottomrule
\end{tabular} 
\end{table}

\section{Experimental Results}\label{sec:experiments}
In this section, we compare CoReS against baselines and the \textit{Backward Compatible Training} (BCT) \cite{DBLP:conf/cvpr/ShenXXS20} state-of-the-art method for different visual search tasks on different benchmark datasets. 
In particular, in Sec.~\ref{sec:cifar_evaluation} we evaluate open-set verification with single and multi-model upgrading on the \cifar-100/10 datasets~\cite{Krizhevsky2009LearningML}; in Sec.~\ref{sec:casia} and Sec. \ref{sec:market}, we analyze single and multi-model upgrading in more challenging tasks, namely face verification (in Sec.~\ref{sec:casia}) on the \webface/LFW datasets~\cite{DBLP:journals/corr/YiLLL14a,LFWTech} and person re-identification (in Sec.~\ref{sec:market}) on the Market1501 dataset~\cite{zheng2015scalable}; 
in Sec.~\ref{sec:gld} and Sec.~\ref{sec:met}, we evaluate CoReS in challenging long-tail distribution datasets, namely Google Landmark Dataset v2 \cite{weyand2020google} and MET dataset \cite{ypsilantis2021met}.
The original datasets are split so that training-sets used for each upgrade have the same number of classes.
Finally, in Sec. \ref{sec:Qualitative Results}, we report a qualitative analysis that provides evidence of how stationarity contributes to  compatibility.
All the representation models were trained using 4 NVIDIA Tesla A-100 GPUs.

\begin{table}[t] 
\caption{Compatibility evaluation for open set verification on \cifar100/10 with two upgrades with CoReS and BCT. Models are learned with with 33\%, and upgraded with 66\% and 100\% of training data. Columns indicate accuracy($M$), whether the Empirical Compatibility Criterion is verified ($ECC$) and the Update Gain ($\Gamma$).}
\label{tab:3step} 
\small
\setlength{\tabcolsep}{6pt}
\sisetup{detect-weight,mode=text,table-format=1.2,round-precision=2,round-mode=places}
\centering 
    \begin{tabular}{c S c c S c c} 
        \toprule
        \multirow{3}{*}{\shortstack{\textsc{Comparison} \\ \textsc{Pair}}} 
        & \multicolumn{3}{c}{\textsc{CoReS}}
        & \multicolumn{3}{c}{\textsc{\textsc{BCT}}}\\ 
         
        \cmidrule(lr){2-4} \cmidrule(lr){5-7} 
        
        & $M$ & \textsc{ECC} & $\Gamma$ (\%) & $M$ & \textsc{ECC} & $\Gamma$ (\%)\\
        \midrule
        $(\phi_{1}, \phi_{1})$ &  0.5853 & \textendash & \textendash & 0.5855 & \textendash & \textendash  \\
        $(\phi_{2}, \phi_{1})$ & \bfseries 0.6093 & $\surd$ & \bfseries 54.7 & 0.5870 & $\surd$ & 5.2  \\
        $(\phi_{3}, \phi_{1})$ & \bfseries 0.5970 & ${\pmb{\pmb{\surd}}}$ & \bfseries 18.5 & 0.5782 & $\times$ & \textendash  \\
        \midrule
        $(\phi_{2}, \phi_{2})$ & \bfseries 0.6292 & \textendash & \textendash & 0.6143 & \textendash & \textendash  \\
        $(\phi_{3}, \phi_{2})$ & \bfseries 0.6139 & $\times$ & \textendash & 0.5876 & $\times$ & \textendash  \\
        \midrule
        $(\phi_{3}, \phi_{3})$ & \bfseries 0.6485 & \textendash & \textendash & 0.6402 & \textendash & \textendash  \\
        \bottomrule
    \end{tabular} 
\end{table}

\subsection{Compatibility Evaluation on \cifar-100/10}
\label{sec:cifar_evaluation}

\begin{figure}[t]
    \subfigure[CoReS]{ \label{fig:5_step_CORES}
        \hspace{-5pt}
        \includegraphics[width=0.40\linewidth]{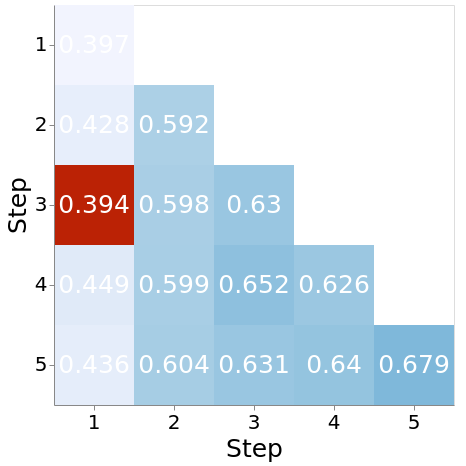}
    }
    \hfill
    \subfigure[BCT~\cite{DBLP:conf/cvpr/ShenXXS20}]{ \label{fig:5_step_BCT}
        \hspace{-2pt}
        \includegraphics[width=0.40\linewidth]{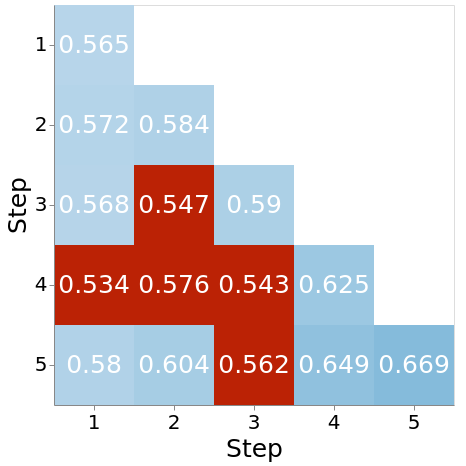}
        \includegraphics[width=0.062\linewidth]{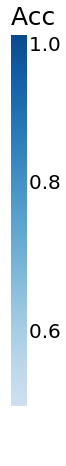}
    }
\caption{Compatibility matrices of  CoReS and BCT for open set verification on \cifar-100/10 with 4-step multi-model upgrading. Diagonal and off-diagonal elements respectively report self-test and cross-test accuracy. Models are sequentially learned on \cifar-100 with 20\%, 40\%, 60\%, 80\%, 100\% of data. Models that do not satisfy the compatibility criterion are highlighted in red.}
\label{fig:cifar_5_step}

\end{figure}
\begin{figure*}[t]
    \centering
    \begin{minipage}{0.76\textwidth}
        
        \subfigure[CoReS]{ \label{fig:10_step_CORES}
            \includegraphics[width=0.492\linewidth]{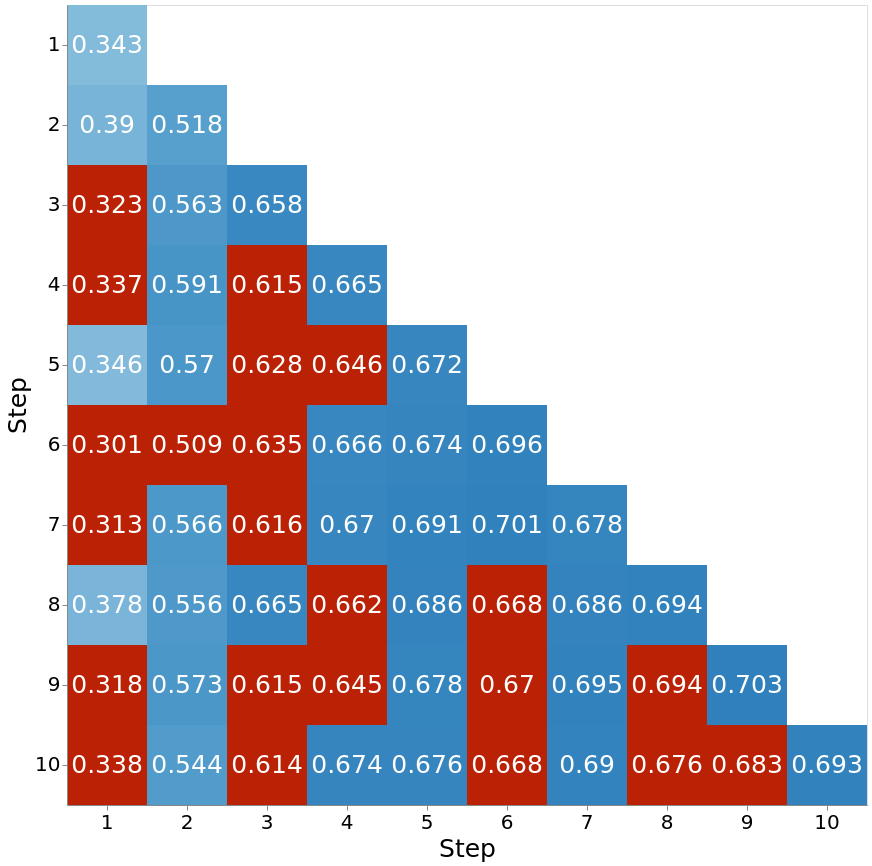}
            
        }
        \subfigure[BCT~\cite{DBLP:conf/cvpr/ShenXXS20}]{ \label{fig:10_step_BCT}
            \includegraphics[width=0.49\linewidth]{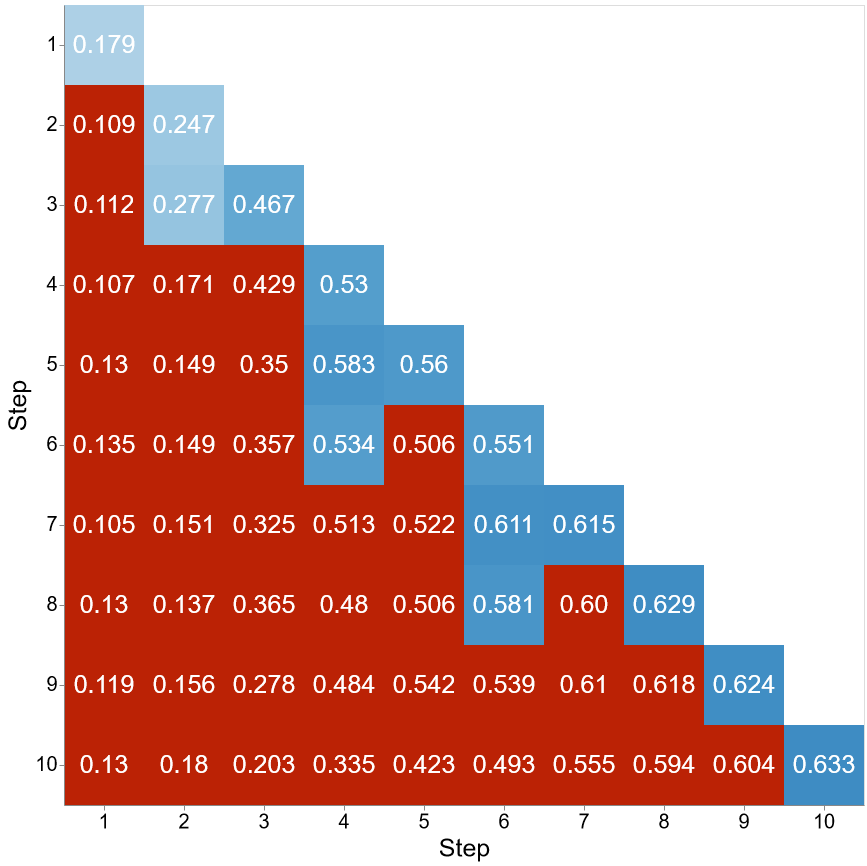}
            \includegraphics[width=0.033\linewidth]{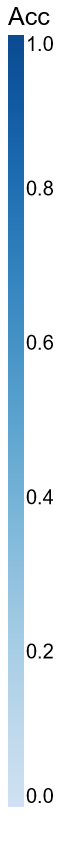}
        }
    \end{minipage}
\caption{Compatibility matrices of CoReS and BCT for open-set  verification on \cifar100/10 with 9-step multi-model upgrading. Diagonal and off-diagonal elements report self-test and cross-test accuracy, respectively. Models are sequentially learned with 10\%, 20\%, ..., 100\% of training data. Models that do not satisfy compatibility are highlighted in red.}
\label{fig:cifar_10_step}

\end{figure*}

We perform open-set verification with \cifar-100/10 datasets. The \cifar-100 and \cifar-10 datasets consist of 60,000 $32 \times 32$ RGB images (50,000 for training and 10,000 for test) in 100 and 10 classes, respectively. Their classes are mutually exclusive. 
We use the \cifar-100 training-set to create the training-sets for upgrading the models and the \cifar-10 test-set for generate the verification pairs used in the open-set verification protocol. We do not use the test-set of \cifar-100 and the training-set of \cifar-10.

The SENet-18 architecture~\cite{senet} adapted to the \cifar $32\times32$ image size\footnote{SENet-18 
code: https://github.com/kuangliu/pytorch-cifar}  is used.
The same random seed is used for initialization every time the model is upgraded.
For CoReS, the output nodes of the $d$-Simplex fixed classifier are set to ${K}_p=100$ so resulting in a feature space is $99$-dimensional. 
Optimization is performed using SGD with $0.1$ learning rate, $0.9$ momentum, and $5 \cdot 10^{-4}$ weight decay. The batch size is set to $128$. 
With every upgrade, training is terminated after $100$ epochs. Learning rate is scheduled to decrease to $0.01$ after $70$ epochs.

We compared CoReS against the $\ell^2$, the \emph{Incremental Fine-Tuning} (IFT), the \textit{Learning without Forgetting} (LwF) \cite{DBLP:conf/eccv/LiH16} baselines and the \textit{Backward Compatible Training}  (BCT) method \cite{DBLP:conf/cvpr/ShenXXS20}.
All the baselines except IFT, are the public implementations of \cite{DBLP:conf/cvpr/ShenXXS20}\footnote{{https://github.com/YantaoShen/openBCT}}.
The $\ell^2$ approach achieves compatibility using an auxiliary loss that takes into account the Euclidean distance $\ell^2$ between features of images in ${\cal T}_{\rm old}$ as evaluated by $\phi_{\rm new\_\ell^2}$ and $\phi_{\rm old}$, that is:
$$
\label{eq:bct_l2}
 \mathcal{L}_{\rm dist} = \frac{1}{\lvert\mathcal{T}_{\rm old}\rvert}\sum_{\stackrel{\mathbf{x} \in  \cal{T}_{\rm old}}{}}
 {\rm dist}(\phi_{\rm new\_\ell^2}(\mathbf{x}),\phi_{\rm old}(\mathbf{x})).
$$
According to this, the new representation model $\phi_{\rm new\_\ell^2}$ is trained as:
$$
\phi_{\rm new\_\ell^2} = \arg\min_{\phi} \Big( \mathcal{L}(\phi, {\cal T}_{\rm new}) + \lambda \mathcal{L}_{\rm dist}(\phi, \phi_{\rm old}, \mathcal{T}_{\rm old}) \Big)
$$
where $\mathcal{L}$ is the standard cross-entropy loss and the scalar $\lambda$ balances the two losses.
The baseline ITM (Independently Trained Models) learns the representation of two versions of models trained independently with standard cross-entropy loss. 
When the gallery is re-indexed, ITM may be thought of as the standard approach to observe, if any, a reduction in performance of a compatibility-based method. The decrease is meant to accommodate the increased requirement that the learning model must fulfill to learn a compatible representation. ITM and CoReS are related since both methods use standard cross-entropy loss and linear classifiers. This relationship allows to quantify the representation capability used for compatibility purposes. Indeed, since the self-test values in the compatibility matrices are calculated with both the query and the gallery images extracted from the same model (i.e., re-indexing the gallery), the difference of the main diagonal entries of ITM and CoReS quantifies the neural network model capacity used to learn compatibility.

To perform open-set verification, we use 3,000 positive and 3,000 negative pairs randomly generated from the \cifar-10 test-set.
Given a pair of images of \cifar-10, verification assesses whether they are of the same class using the cosine distance.
One of the images of each pair can be considered gallery and the other the query.
Compatibility is evaluated for one, two, four and nine upgrade steps.

In the one-upgrade case, the model is trained with 50\% of \cifar-100 classes and upgraded using 100\%. 
The comparative evaluation is shown in Tab.~\ref{tab:baselinecomp} in terms of verification accuracy ($M$), whether the Empirical Compatibility Criterion is verified (ECC) and  Update Gain ($\Gamma$). 
It can be noticed that only BCT and CoReS satisfy Eq.~\ref{eq:compatible_set}. However, CoReS obtains higher Update Gain with respect to BCT. 

In the two-upgrade case, the model is initially trained on 33\% of \cifar-100 classes and upgraded with 66\% and 100\%. 
The compatibility matrix is shown in Fig. \ref{fig:multi_step_with_baseline} for all the methods compared.
We can notice that none of the methods achieves full compatibility in this case, however BCT and CoReS achieve higher accuracy values with respect to the others. CoReS misses compatibility only once between $\phi_{3}$ and $\phi_{2}$.
In Tab.~\ref{tab:3step} the behavior of CoReS and BCT in the  two-upgrade  case is reported in more detail.
We can observe that CoReS has a substantially higher compatibility and update gain than BCT.
This may be due to the fact that BCT learns compatibility of $\phi_3$ with no direct relationship with $\phi_1$, while CoReS learns $\phi_3$ taking into account both $\phi_1$ and $\phi_2$, because all models share the same class prototypes in the feature space.

The compatibility matrices of CoReS and BCT for the cases of four and nine upgrade steps are shown in Fig. \ref{fig:cifar_5_step} and \ref{fig:cifar_10_step}, respectively. 
The superior behavior of CoReS with respect to BCT is clearly evident in both cases, scoring $0.9$ versus $0.5$ and $0.51$ versus $0.11$ of \normnummecc, respectively. This indicates that the more upgrades are made, the better CoReS performs with respect to BCT.

\subsection{Compatibility Evaluation on CASIA-WebFace/LFW}
\label{sec:casia}

\begin{table}[t]
    \centering
    \small
    \caption{Compatibility of CoReS and BCT for open-set face verification on the \webface /LFW dataset with one upgrade. Models are learned on \webface with 50\% and upgraded with 100\% of data. Columns indicate accuracy ($M$), whether the Empirical Compatibility Criterion is verified (ECC) and the Update Gain ($\Gamma$).} 
    \label{tab:face_rec}
    \setlength{\tabcolsep}{6pt}
    \begin{tabular}{ccccccc} 
        \toprule 
        \multirow{3}{*}{\shortstack{\textsc{Comparison} \\ \textsc{Pair}}} 
        & \multicolumn{3}{c}{\textsc{CoReS}}
        & \multicolumn{3}{c}{\textsc{\textsc{BCT}}}\\ 
         
        \cmidrule(lr){2-4} \cmidrule(lr){5-7}         
        & $M$ & \textsc{ECC} & $\Gamma$ (\%) & $M$ & \textsc{ECC} & $\Gamma$ (\%) \\

        \midrule
        $(\phi_{\rm old}, \phi_{\rm old})$  & 0.90 & \textendash  & \textendash  & \bfseries 0.91 & \textendash  & \textendash   \\
        $(\phi_{\rm new}, \phi_{\rm old})$  & 0.91 & $\surd$ & \bfseries 0.32 & 0.91 & $\surd$ & 0.29  \\
        $(\phi_{\rm new}, \phi_{\rm new})$  & \bfseries 0.92 & \textendash  & \textendash  & 0.91 & \textendash  & \textendash   \\
        \bottomrule
    \end{tabular} 
\end{table}
\begin{figure}[t]
    \centering
    \subfigure[]{ \label{fig:MECC_face}
        \hspace{-18pt}
        \includegraphics[width=1.05\linewidth]{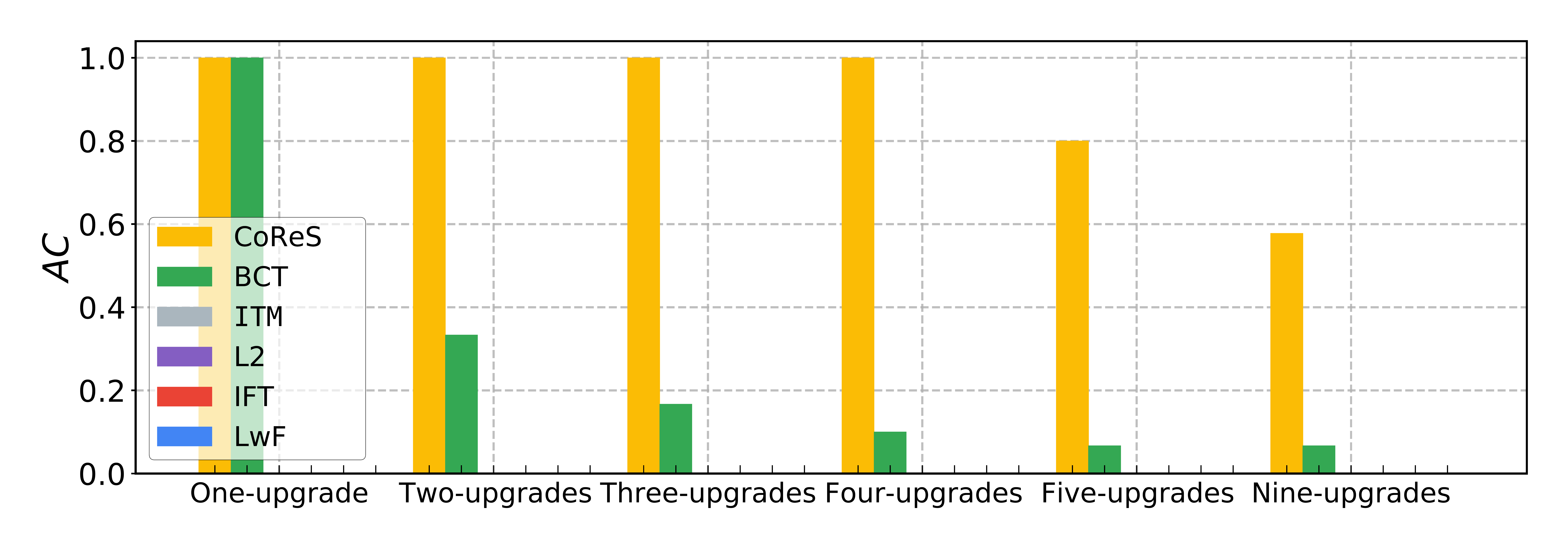}
    }
    \\
    \subfigure[]{ \label{fig:casia_AM}
        \hspace{-18pt}
        \includegraphics[width=1.05\linewidth]{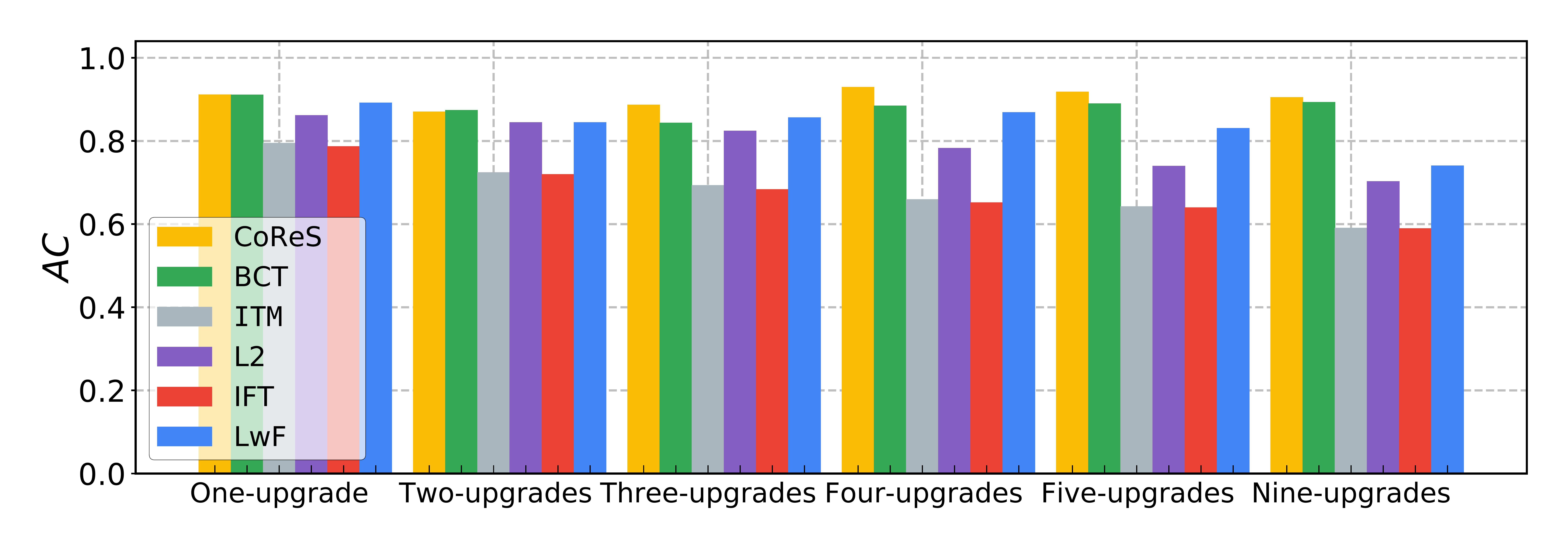}
    }
    \caption{Compatibility of CoReS and compared methods (shown color-coded) for open-set face verification on the \webface/LFW dataset with multi-model upgrading. Bins show:
    (a) \normnummecc scores for different number of upgrades; (b) \avgmetr scores for different number of upgrades.
    }\label{fig:AC-AM-face-withbaseline}
\end{figure}

\begin{figure*}[t]
\centering
    \begin{minipage}{0.9\textwidth}
    \centering
        \subfigure[{ITM}]{\label{fig:face-5step_ub}
            \includegraphics[width=0.22\linewidth]{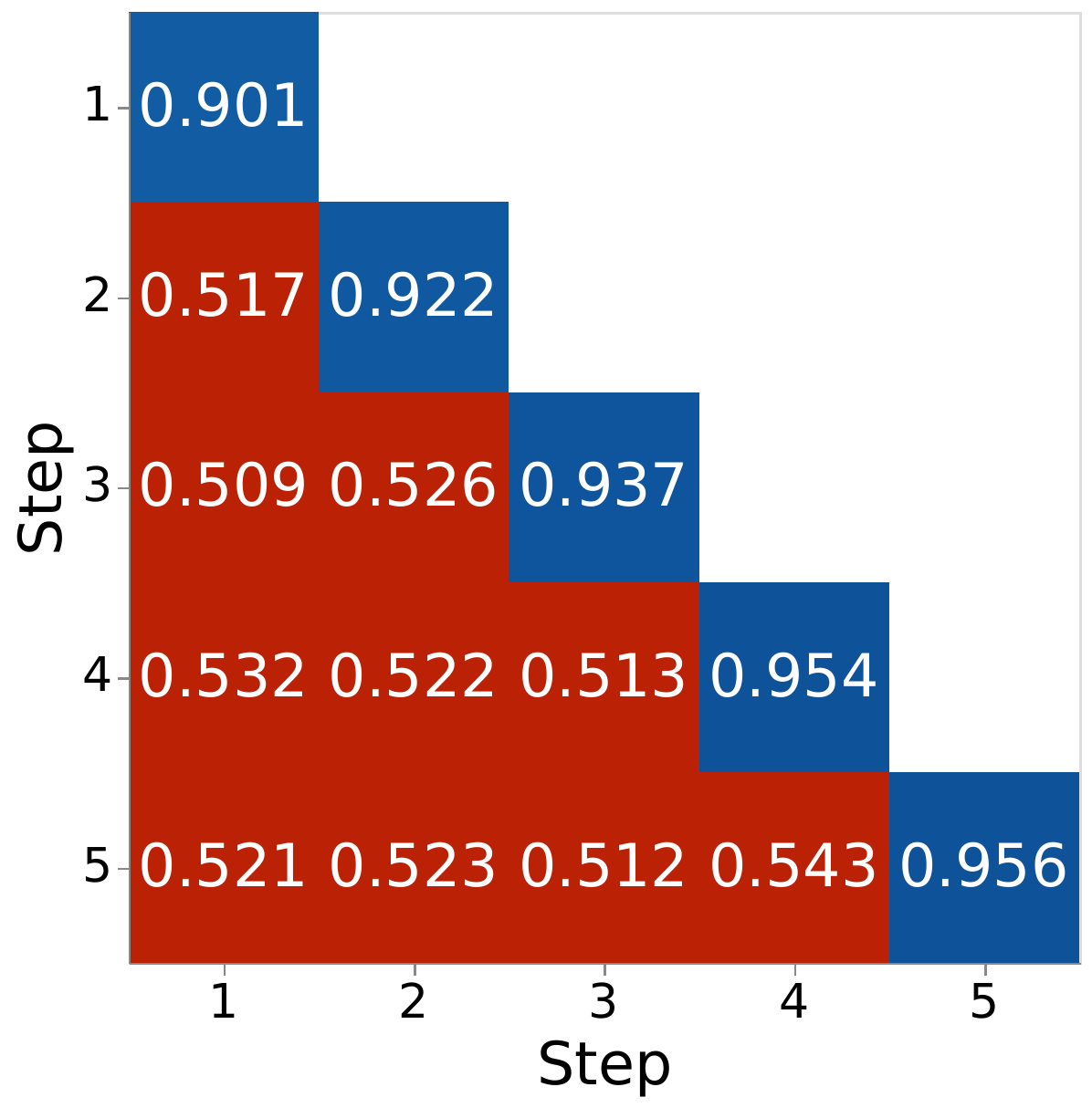}
        }
        \hspace{15pt}
        \subfigure[{$\ell^2$}]{\label{fig:face-5step_l2}
            \includegraphics[width=0.22\linewidth]{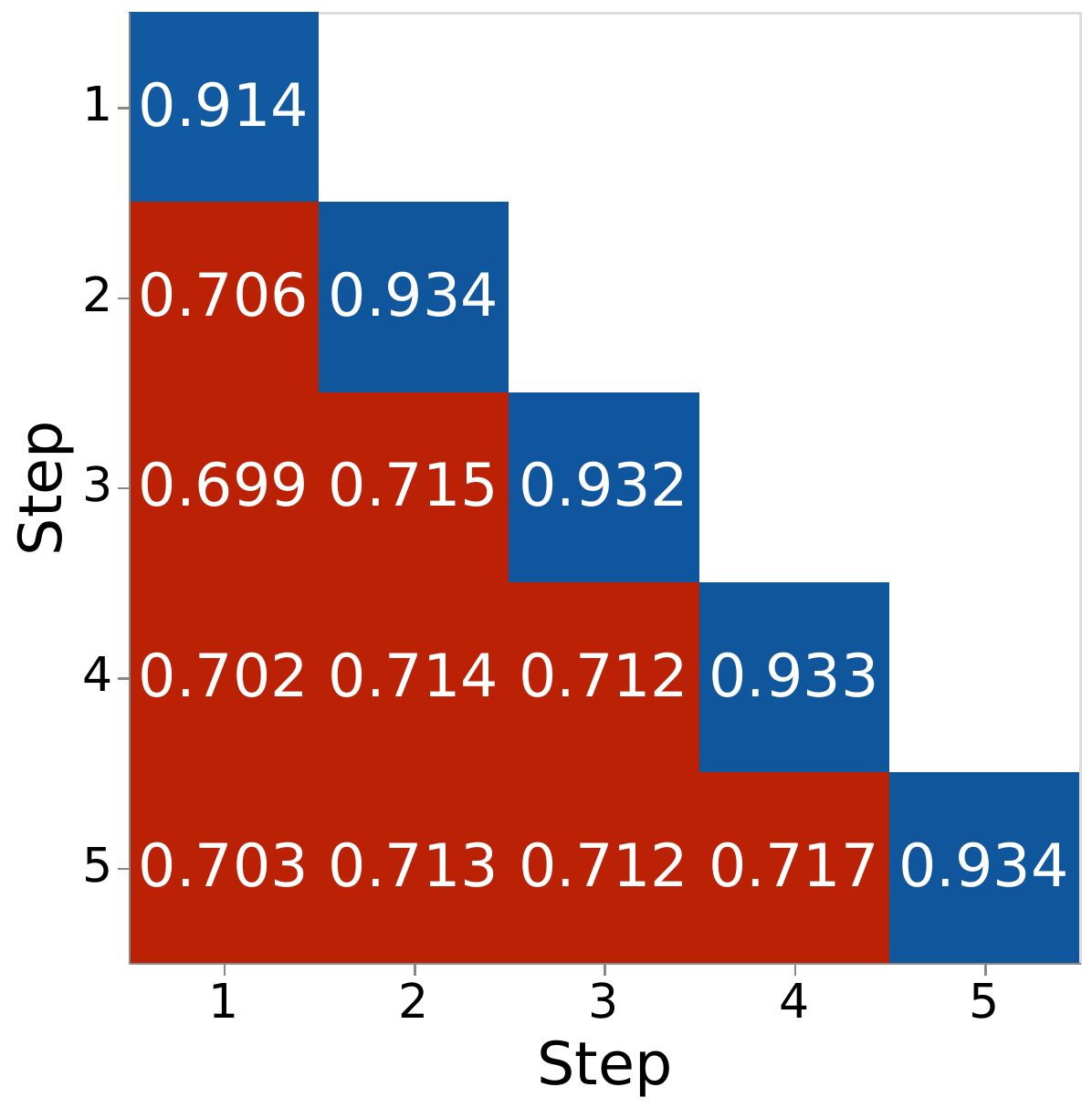}
        }
        \hspace{15pt}
        \subfigure[{IFT}]{ \label{fig:face-5step_ift}
            \includegraphics[width=0.22\linewidth]{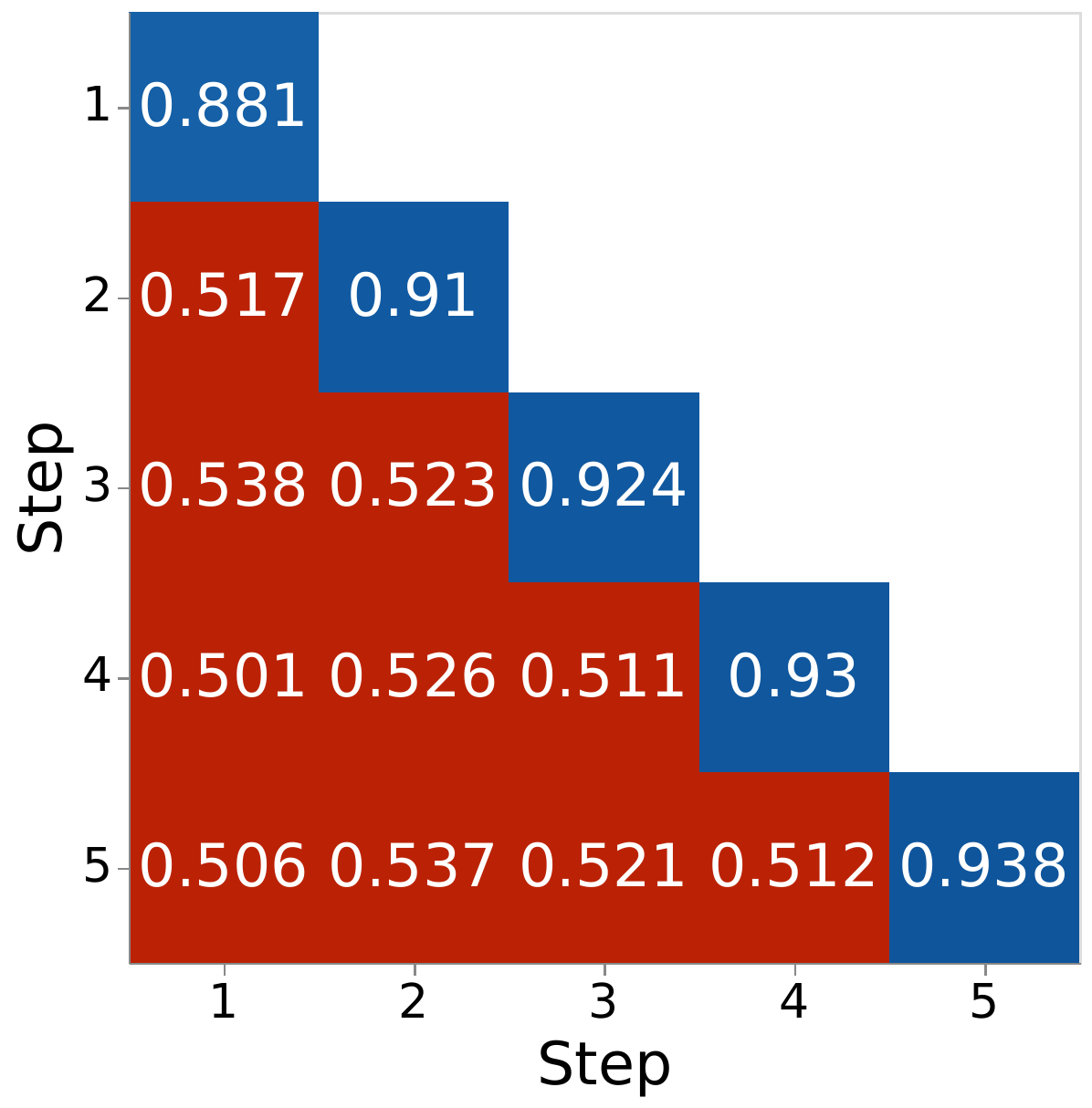}
        }
        \hspace{-5pt}
        \subfigure{
            \subfigcapmargin = 0pt
            \includegraphics[width=0.034\linewidth]{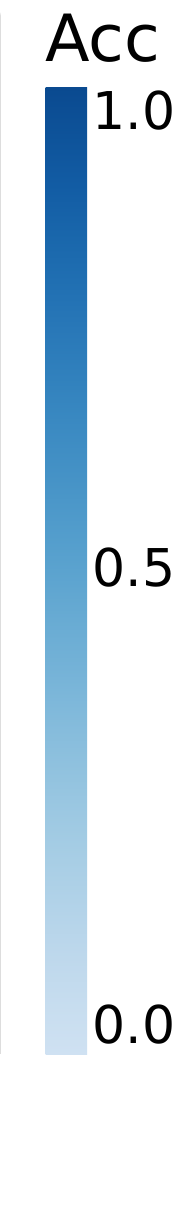}
            \addtocounter{subfigure}{-1}
        }
        \\
        \subfigure[{LwF}]{ \label{fig:face-5step_lwf}
            \includegraphics[width=0.22\linewidth]{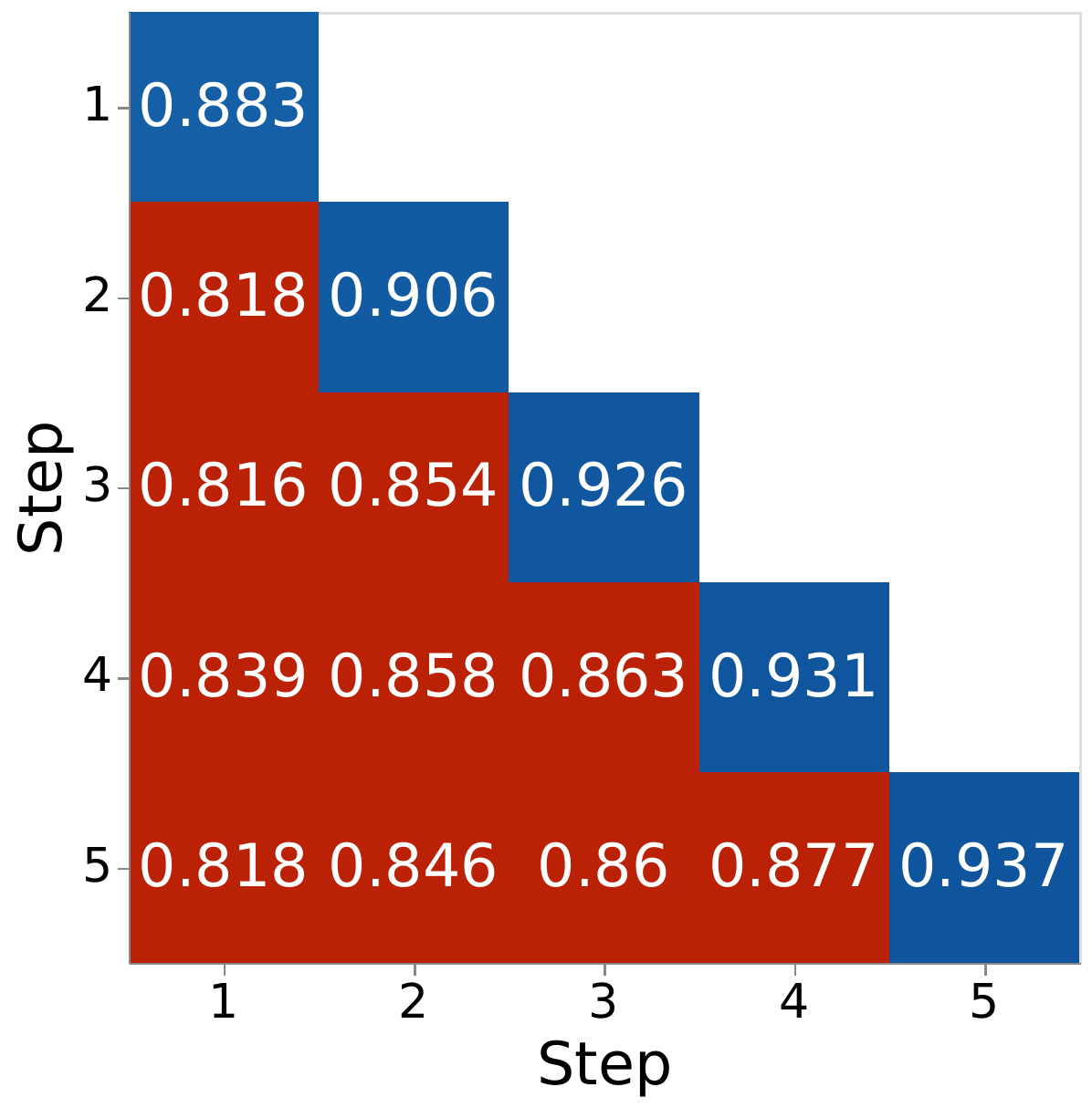}
        }
        \hspace{15pt}
        \subfigure[BCT]{ \label{fig:face-5step_bct}
            \includegraphics[width=0.22\linewidth]{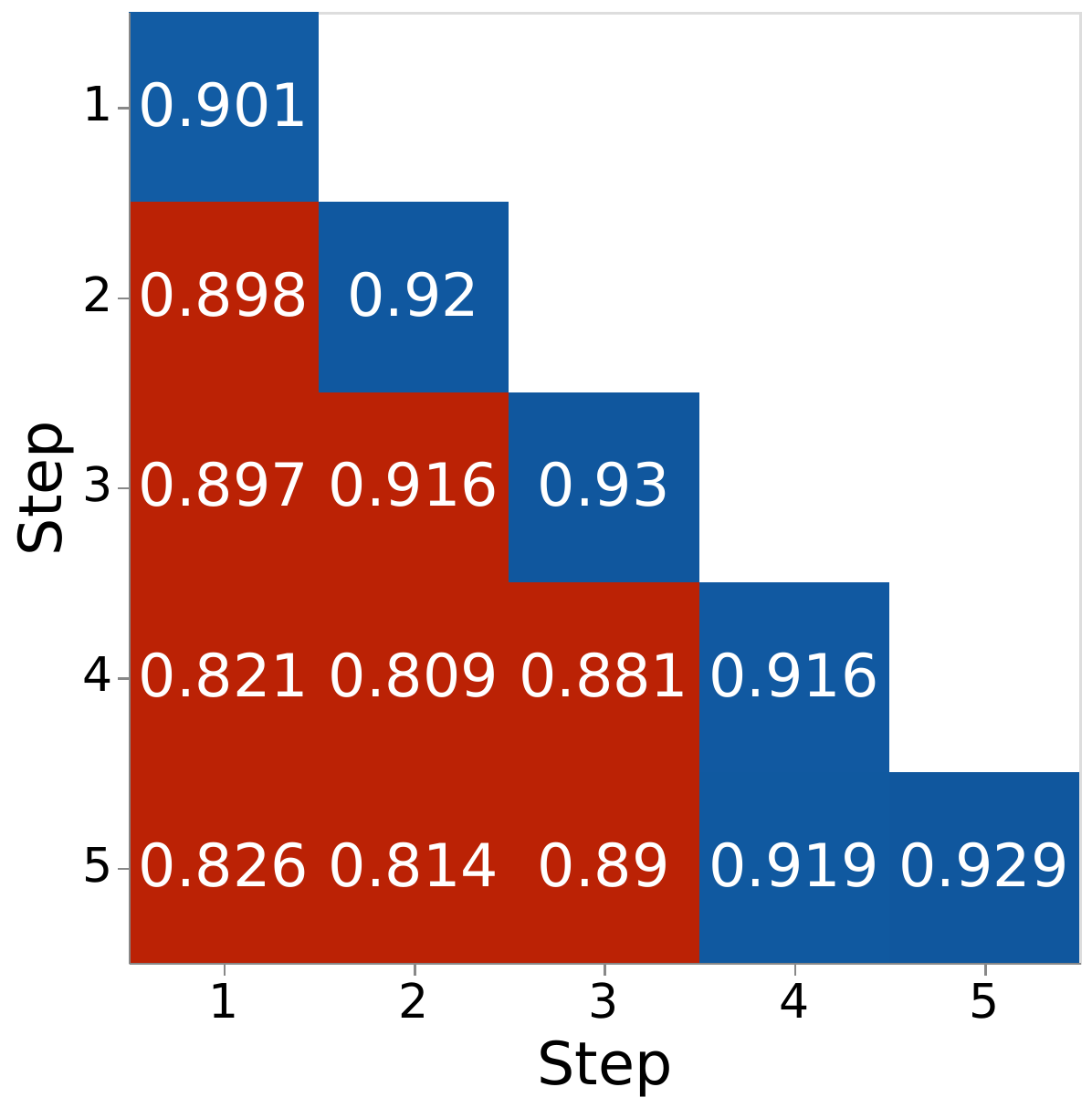}
        }
        \hspace{15pt}
        \subfigure[CoReS]{ \label{fig:face-5step_cores}
            \includegraphics[width=0.22\linewidth]{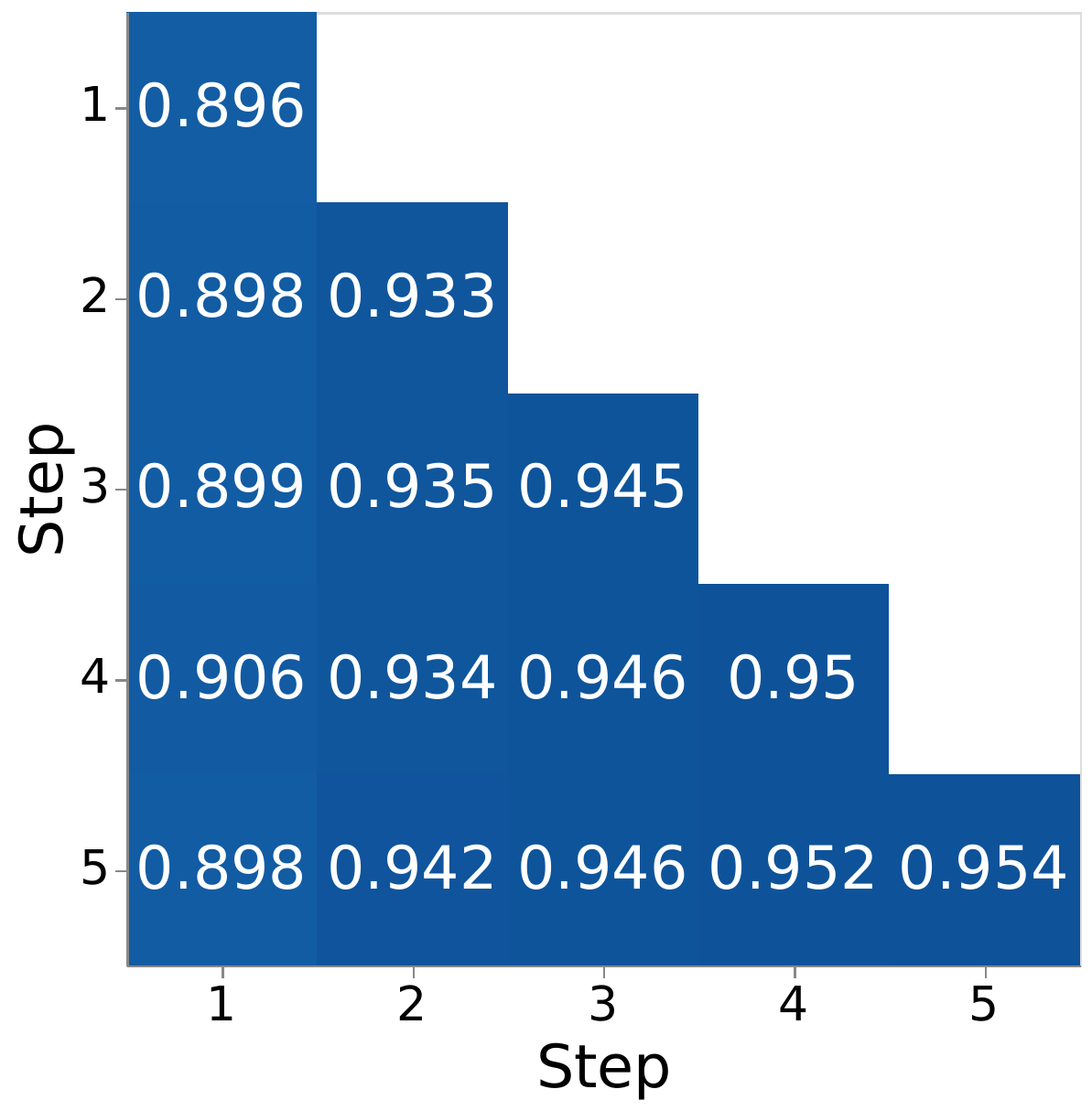}
        }
        \hspace{-5pt}
        \subfigure{
            \subfigcapmargin = 0pt
            \includegraphics[width=0.034\linewidth]{imgs/5step/bar.pdf}
            \addtocounter{subfigure}{-1}
        }
    \end{minipage}
\caption{{Compatibility matrices of CoReS and compared methods for open-set verification on \webface-LFW dataset with 4-step multi-model upgrading. Models are sequentially learned on \webface with 20\%, 40\%, 60\%, 80\%, 100\%. Models that do not satisfy compatibility are highlighted in red.} }
\label{fig:face-5step}

\end{figure*}
In this evaluation, we perform open-set face verification using \webface dataset to create the training-sets and LFW as the test set.
The \webface dataset includes $494,414$ RGB face images of $10,575$ subjects. The LFW dataset contains $13,233$ target face images of $5,749$ subjects. Of these, $1,680$ have two or more images, while the remaining $4,069$ have only one single image. 
ResNet50~\cite{DBLP:conf/cvpr/HeZRS16} with input size of $112 \times 112$ is used as backbone. 
Optimization is performed using SGD with $0.1$ learning rate, $0.9$ momentum, and $5 \cdot 10^{-4}$ weight decay. The batch size is $1,024$. With every upgrade, training is terminated after 120 epochs. Learning rate is scheduled to decrease to $0.01$, $0.001$, and $0.0001$ at epoch $30$, $60$, $90$ respectively.

Compatibility is evaluated for one, two, three, four, five, and nine upgrade steps.

In the one-upgrade case, models are learned with 50\% of the \webface dataset and upgraded with 100\%. 
Tab.~\ref{tab:face_rec} shows that, with 50\% of \webface dataset, CoReS and BCT achieve similar performance. This is because there is already sufficient data variability to learn compatible features. 

A big difference between CoReS and compared methods becomes evident when multiple upgrades are considered as shown in Fig.~\ref{fig:MECC_face}. 
In this figure, the values of the \normnummecc are reported for each method over a set of different experiment respectively with one, two, three, four, five, and nine upgrades.
CoReS achieves full compatibility ($AC = 1$) for one, two, three, and four upgrades and starts decreasing \normnummecc from five upgrade steps up to $AC = 0.58$ with nine upgrades. 
In contrast BCT loses performance already with two upgrade steps finishing at $AC = 0.09$ with nine upgrade steps.
Baseline methods report $AC = 0$ in all of the scenario.
The compatibility matrices for the case of four upgrades are shown in detail in Fig. \ref{fig:face-5step_cores}. In this case, we can appreciate full compatibility of CoReS with respect to the poor compatibility of BCT and other methods. 
In Fig.~\ref{fig:casia_AM}, we report the \avgmetr metric for these experiments.
It can be observed that the CoReS and BCT score almost the same average verification accuracy in all the experiments, while in the others values are always lower. This is due to the fact that cross-test values are low since no compatibility is reported.

We conclude that for multi-model upgrading, CoReS, while having the same verification performance of BCT, largely improves compatibility across model upgrades with 544\% relative improvement over BCT for the challenging scenario of nine-step upgrading. The lower \normnummecc of BCT appears to be related to the fact that in this method compatibility is obtained only through transitivity from the model previously learned.

\subsection{Compatibility Evaluation on Market1501} \label{sec:market}

\begin{table}[t]
\small
\centering
\caption{Compatibility evaluation for person re-identification on the Market1501 dataset for one and two-step upgrading. 
\normnummecc and \avgmetr are used to compare CoReS with other methods. mAP is used as \\evaluation metric $M$.}
\label{tab:market} 
\begin{tabular}{c cccc}
\toprule
\multirow{3}{*}{\shortstack{\textsc{Method}}}                                 & \multicolumn{2}{c}{One-upgrade} & \multicolumn{2}{c}{Two-upgrades} \\
\cmidrule(lr){2-3} \cmidrule(lr){4-5} 
                              & \normnummecc    &  $AM$          & \normnummecc  &  $AM$         \\
\midrule
\textsc{Standard}                                                       & 0     &  {0.30}        & {0}   &   {0.22}     \\
{$\ell^2$} & {0}     &  {0.36}        & {0}   &   {0.31} \\
{\textsc{IFT}}                                                        & {0}   &  {0.30} & {0}   &   {0.21}   \\
{\textsc{LwF} }                                                       &   {0}    &  {0.40}      &    {0}   &    {0.36}            \\
\textsc{BCT}                                                      & 1      & 0.49      & 0.3      & 0.50    \\
\textsc{CoReS}                                                     & 1      & \textbf{0.57}      &  \textbf{1}      & \textbf{0.51}    \\
\bottomrule
\end{tabular}
\end{table}

In this experiment, we perform person re-identification (1:N search) using the Market1501 dataset.
Differently from \webface/LFW, Market1501 includes images of identities with severe occlusions and pose variations. This makes learning compatible features largely more challenging. The Market1501 dataset contains $1,501$ identities, split in $751$ for training and $750$ for test. The two sets do not share identities. Images of each identity are captured by several cameras so that cross-camera search can be performed. For 1:N re-identification, a set of templates is used in the gallery-set. Then templates of the query-set are used to search against the gallery templates. Mean Average Precision (mAP) is used as quality metric.
Following~\cite{torchreid}, we use a pre-trained ResNet101~\cite{DBLP:conf/cvpr/HeZRS16} as backbone and Adam~\cite{kingma2014adam} as optimizer with an initial learning rate of $3\cdot10^{-4}$.
The batch size is 256.
With every upgrade, training is terminated after 25 epochs. Learning rate is scheduled to decrease to $3\cdot10^{-5}$ after 20 epochs.

Tab.~\ref{tab:market} shows the results for one and two upgrade steps.
We observe that CoReS always achieves full compatibility, while BCT loses compatibility already with two upgrade steps. 
The baseline methods never achieve compatibility in both one and two-step upgrade.
While it is evident that the stationarity of the CoReS representation still plays a key role to ensure compatibility across the upgrades also in this challenging task, the results confirm 
the limits of transitive compatibility used by BCT.

\subsection{Compatibility Evaluation on GLDv2} \label{sec:gld}
Google Landmark Dataset v2 (GLDv2), cleaner version, consists of $81,313$ classes and $1,580,470$ images \cite{weyand2020google}. 
The main issue is that with such a large number of classes, the $d$-Simplex classifier matrix will be large and will require additional GPU memory. Although we have already evaluated the fixed classifier by training with more than 10k classes using \webface dataset, with 81k classes it is necessary to also reduce the representation dimension. This is a result of the linear growth of the feature dimension of the $d$-Simplex fixed classifier with the number of classes. However, training can still be achieved effectively because fixed weights do not require additional storage for the gradient, thus resulting in a decrease in the amount of memory needed. 
The challenging situation remains at test time in which the gallery-set is composed of more than 760k elements, each of which is $81,312$ dimensional, requiring more than 200Gb of storage. With such a memory requirement, the Faiss GPU accelerated indexing \cite{johnson2019billion} available in the GLDv2 repository\footnote{\url{https://github.com/cvdfoundation/google-landmark}} may not be fully suitable for use. To overcome this issue, inspired by \cite{lin2018dgc}, we adopted top-$k$ sparsification\footnote{All the entries of a feature representation vector are set to zero except for the top $k$.} and perform nearest neighbor using the PySparNN library\footnote{\url{https://github.com/facebookresearch/pysparnn}}. 
Tab.~\ref{tab:gld} shows the retrieval results on GLDv2 with one-step upgrading for both CoReS and BCT. In our experiments, we set the top-$k$ sparsification to $k = 512$, use the ResNet50 backbone and image size $256 \times 256$. As it can be seen from the table, CoReS achieves compatibility also under this learning settings, while BCT results not compatible as the cross-test is $0.16$ which is lower than the previous self-test which results $0.18$. 
CoReS performs better than BCT in learning compatible representation with a long-tail distribution dataset. One possible reason is that CoReS is based on a $d$-Simplex fixed classifier that has recently been shown to perform better than a learnable classifier in long-tail/imbalanced datasets \cite{yang2022we,kasarla2022maximum}.
\begin{table}[t]
    \centering
    \small
    \caption{Compatibility of CoReS and BCT on the retrieval setup of GLDv2 with 1-step upgrade. Models are learned on GLDv2 with 50\% of classes and upgraded with 100\% of data. Columns indicate mAP@100 ($M$), whether the Empirical Compatibility Criterion is verified (ECC) and the Update Gain ($\Gamma$).}
    \label{tab:gld}
    \setlength{\tabcolsep}{6pt}
    \begin{tabular}{ccccccc} 
        \toprule 
        \multirow{3}{*}{\shortstack{\textsc{Comparison} \\ \textsc{Pair}}} 
        & \multicolumn{3}{c}{\textsc{CoReS}}
        & \multicolumn{3}{c}{\textsc{\textsc{BCT}}}\\ 
         
        \cmidrule(lr){2-4} \cmidrule(lr){5-7}         
        & $M$ & \textsc{ECC} & $\Gamma$ (\%) & $M$ & \textsc{ECC} & $\Gamma$ (\%) \\
        
        \midrule
        $(\phi_{\rm old}, \phi_{\rm old})$  & \bfseries 0.19 & \textendash  & \textendash  & 0.18 & \textendash  & \textendash   \\
        $(\phi_{\rm new}, \phi_{\rm old})$  & \bfseries 0.20 & $\surd$ & \bfseries 0.15 & 0.16 & $\times$ & \textendash  \\
        $(\phi_{\rm new}, \phi_{\rm new})$  & \bfseries 0.21 & \textendash  & \textendash  & 0.19 & \textendash  & \textendash   \\
        \bottomrule
    \end{tabular} 
\end{table}

\subsection{Compatibility Evaluation on MET} \label{sec:met}
MET, mini version, is an instance-level artwork recognition dataset consisting of $38,307$ images and $33,501$ classes \cite{ypsilantis2021met}.
The main issue for this dataset is that standard supervised learning with cross-entropy is not converging. The official repository\footnote{ \url{https://github.com/nikosips/met}} provides a self-supervised training baseline based on contrastive learning \cite{chopra2005learning} which, starting from a pre-trained model on ImageNet, converges to a working representation. 
We have extended CoReS to support the contrastive learning setup. 
The extension consists in using the output logits of the $d$-Simplex fixed classifier as the contrasted representation. 
Since classes are not required in contrastive learning, under this condition the dimension $d$ of the fixed classifier becomes a free parameter (we set $d=2048$). 
Our training in miniMET starts by pre-training a $d$-Simplex fixed classifier based ResNet50 on ImageNet as in \cite{perniciTNNLS2021}. Contrastive learning with CoReS is then accomplished by using a distinct $d$-Simplex fixed final matrix which operates as a projector rather than a classifier. 
MiniMET is used to create the training-sets for the upgrade and the gallery-set, while the MET dataset of the queries is used as query-set. 
Tab.~\ref{tab:met} shows the result with one-step upgrading of CoReS and BCT. As evidenced by the table, only CoReS achieves compatibility. This finding further supports the positive performance of the proposed method, and suggests that it may also be used for contrastive/self-supervised learning. The positive results may be motivated by the fact that the extension of CoReS to contrastive learning is related to DirectCLR \cite{jing2021understanding} in which the SimCLR contrastive method \cite{chen2020simple} is trained with a final fixed projector based on random weights. 
We can consider the $d$-Simplex classifier in this ``contrastive extended CoReS'' as a special fixed projector. The positive results we achieved are consistent with the improved results on SimCLR reported by DirectCLR.
\begin{table}[t]
    \centering
    \small
    \caption{\textcolor{black}{Compatibility of CoReS and BCT on the kNN classification of MET with 1-step upgrade. Models are learned on miniMET with 50\% of classes and upgraded with 100\% of data. Columns indicate accuracy ($M$) and whether the Empirical Compatibility Criterion is verified (ECC) and the Update Gain ($\Gamma$).} } 
    \label{tab:met}
    \setlength{\tabcolsep}{6pt}
    \begin{tabular}{ccccccc} 
        \toprule 
        \multirow{3}{*}{\shortstack{\textsc{Comparison} \\ \textsc{Pair}}} 
        & \multicolumn{3}{c}{\textsc{CoReS}}
        & \multicolumn{3}{c}{\textsc{\textsc{BCT}}}\\ 
         
        \cmidrule(lr){2-4} \cmidrule(lr){5-7}         
        & $M$ & \textsc{ECC} & $\Gamma$ (\%) & $M$ & \textsc{ECC} & $\Gamma$ (\%) \\
        
        \midrule
        $(\phi_{\rm old}, \phi_{\rm old})$  & 0.59 & \textendash  & \textendash  & \bfseries 0.63 & \textendash  & \textendash   \\
        $(\phi_{\rm new}, \phi_{\rm old})$  & 0.65 & $\surd$ & \bfseries 0.6  & 0.54 & $\times$ & \textendash  \\
        $(\phi_{\rm new}, \phi_{\rm new})$  & \bfseries 0.69 & \textendash  & \textendash  & 0.68 & \textendash  & \textendash   \\
        \bottomrule
    \end{tabular} 
\end{table}

\subsection{Qualitative Results}
\label{sec:Qualitative Results}

\begin{figure}[t]
    \subfigure[\textsc{IFT}]{ \label{fig:ft_spc_ift}
        \hspace{-0.4cm}
        \adjincludegraphics[width=1.0\linewidth,trim={0 {.67\height} 0 {.0\height}},clip]{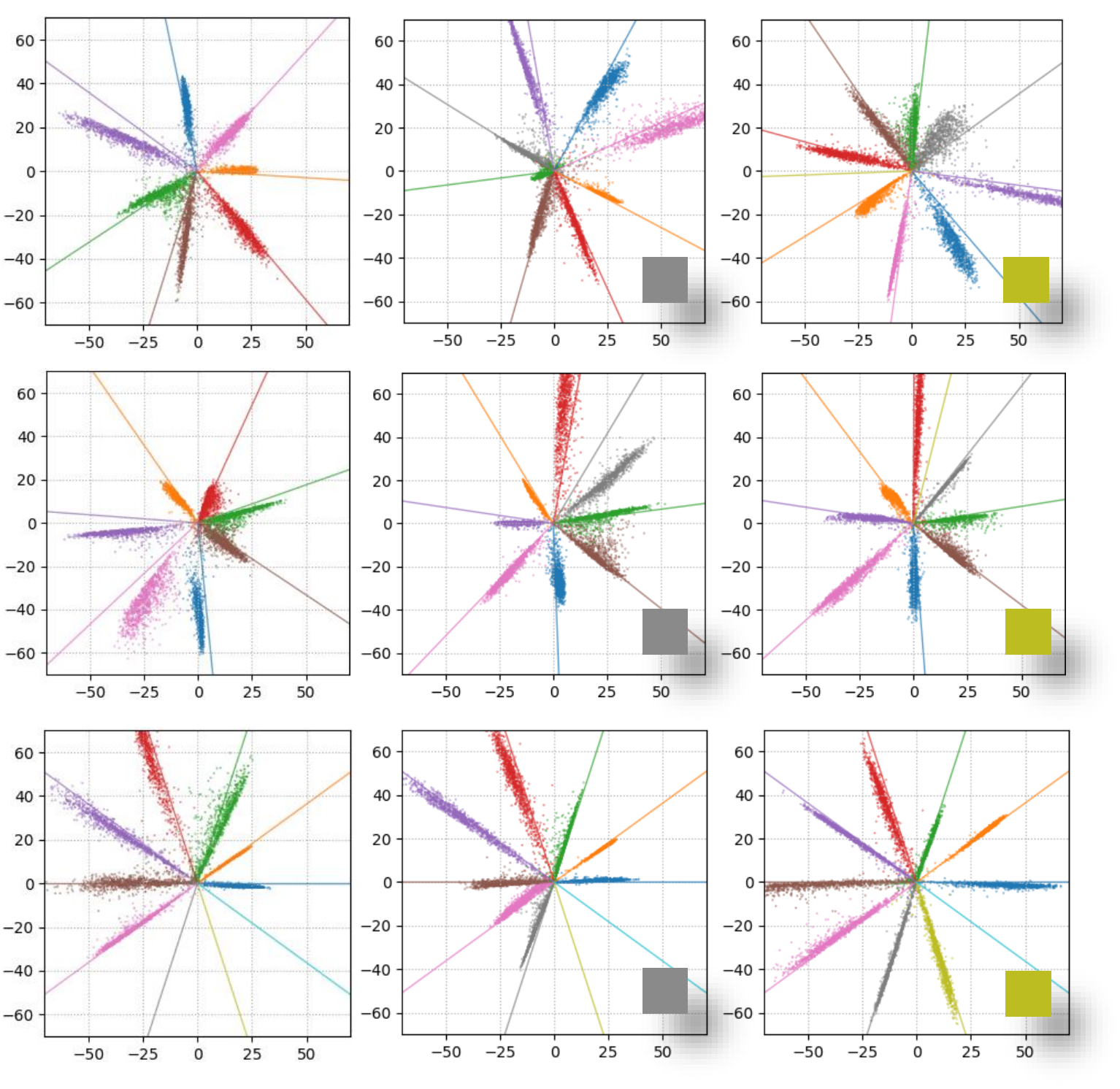}
    }
    \subfigure[\textsc{BCT}~\cite{DBLP:conf/cvpr/ShenXXS20}]{ \label{fig:ft_spc_bct}
        \hspace{-0.4cm}
        \adjincludegraphics[width=1.0\linewidth,trim={0 {.34\height} 0 {.33\height}},clip]{imgs/feat/2d_feat_comparison.pdf}
    }
    \subfigure[\textsc{CoReS}]{ \label{fig:ft_spc_cores}
        \hspace{-0.4cm}
        \adjincludegraphics[width=1.0\linewidth,trim={0 0 0 {.66\height}},clip]{imgs/feat/2d_feat_comparison.pdf}
    }
    \caption{
    Feature spatial configuration with 2-step upgrading (1 class per upgrade) and MNIST dataset with 2D features. Colored cloud points are the features from the test-set and lines represent the classifier prototypes. Compatibility methods compared: (a) IFT; (b) BCT; (c) CoReS. 
    \vspace{-0.3cm}
    }
    \label{fig:evolution_featu_space}
\end{figure}

Fig.~\ref{fig:evolution_featu_space} shows the different behaviors of IFT, BCT, and CoReS for a simple two-step upgrading with the MNIST dataset.
In this experiment, the training set is upgraded from seven to eight classes and then to nine classes using the same setting as in Sec.~\ref{sec:proposed-stationarity}.

It can be observed that 
IFT changes the spatial configuration of the representation as every new class is learned.
In fact, IFT has no mechanism to prevent the rearrangement of the features when the model is upgraded.
By exploiting the class prototypes of the classifier at the previous upgrade step, BCT keeps the representation reasonably stationary, although small changes are introduced to accommodate the new classes.  
Differently from the others, the features learned by CoReS remain aligned with the class prototypes.
The configuration of the feature space does not change, thus providing compatibility with previous representations.

\section{Ablation Studies}
\label{sec:ablation_studies}
As CoReS is a single building block method, ablation study consists in tuning training hyperparameters. In the following, we present experiments to evaluate the factors that affect the performance of our training procedure. In particular: in Sec.~\ref{sec:num_epochs}, we discuss how much the number of training epochs affects learning compatible representations; in Sec.~\ref{sec:model_selection}, we evaluate the influence of Model Selection over compatibility; in Sec.~\ref{sec:pre_train}, we analyze the effects of starting from a pre-trained model; in Sec.~\ref{sec:Different Model Initialization}, we examine model initialization, whether using same and different random initialization or fine-tuning from a previously learned model; in Sec.~\ref{sec:different_order_classes}, we consider the effect of using different class sequence ordering; 
in Sec.~\ref{sec:Different Model Architecture} we study how much different network architectures impact compatibility; in Sec.~\ref{sec:diff-numb-future-classes}, we examine whether the number of pre-allocated future classes in the $d$-Simplex classifier impacts the performance of CoReS. Finally, in Sec.~\ref{sec:diff-source-upgrade-data}, we explore whether different source of upgrade data influences the compatibility performance. 

The ablation experiments are performed on the verification task of Sec. \ref{sec:cifar_evaluation} using  \cifar-100/10 dataset with nine upgrades. 

\subsection{Number of Epochs}
\label{sec:num_epochs}

In this experiment, models were learned sequentially with 30, 70, 100, 200 and 350 epochs at each upgrade with 9-step upgrading. We modified the learning rate as follows: no change for the 30-epochs case; scheduled to decrease to 0.1 at epoch 50 for the 70-epochs case; scheduled to decrease to 0.01 at epoch 70 for the 100-epochs case (already discussed in Sec. \ref{sec:cifar_evaluation}); scheduled to decrease to 0.1 at epoch 70 and 0.01 at epoch 140 for the 200 epochs case; scheduled to decrease to 0.1 at epoch 150 and to 0.01 at epoch 250 for the 350 epochs case. 

\begin{figure}[t]
    \hspace{-0.4cm}
    \subfigure[]{ \label{fig:MECC_at_epoch}
        \includegraphics[width=0.487\linewidth]{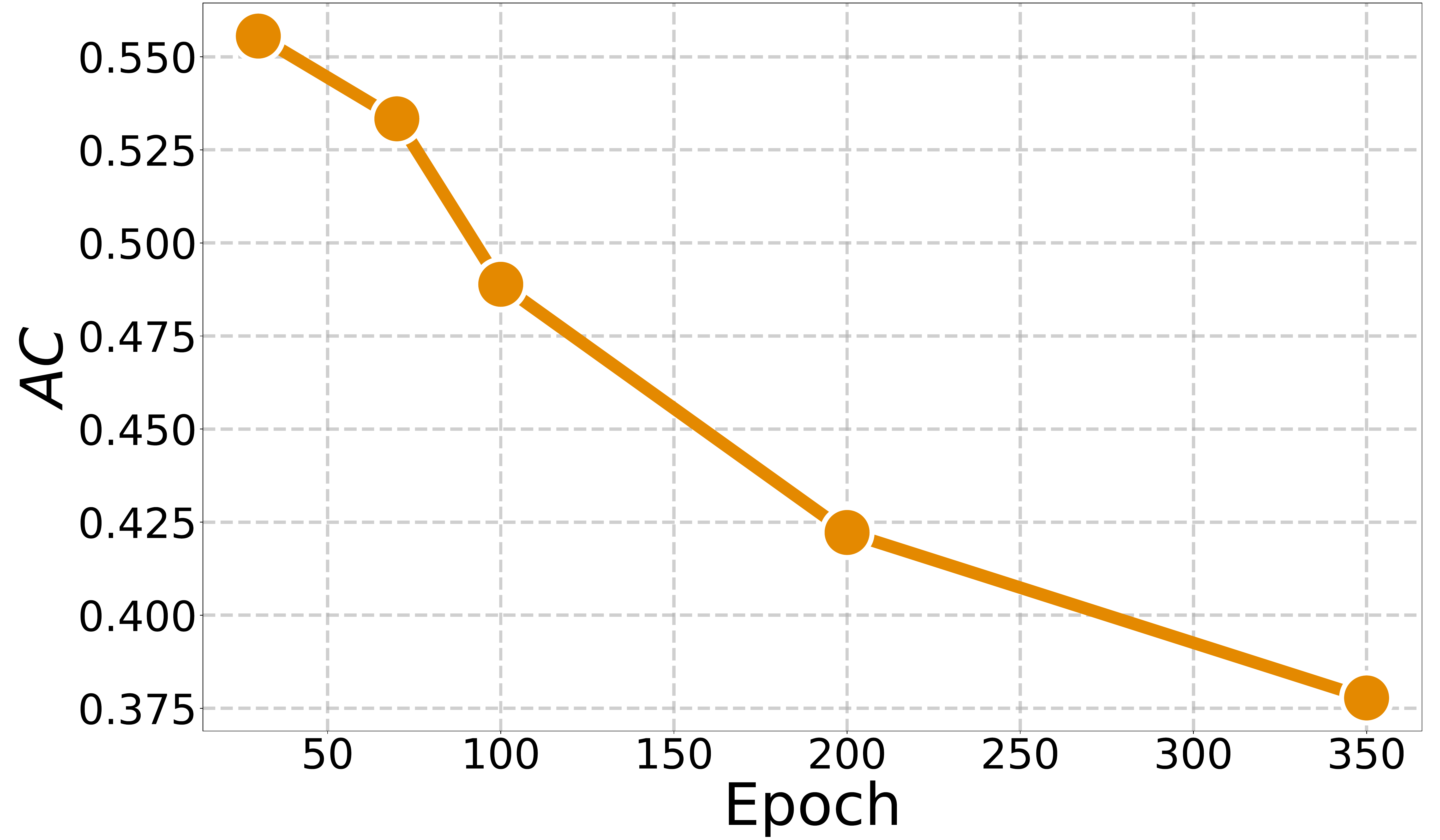}
    }
    \hspace{-0.3cm} 
    \subfigure[]{  \label{fig:M_avg_at_epoch}
        \includegraphics[width=0.487\linewidth]{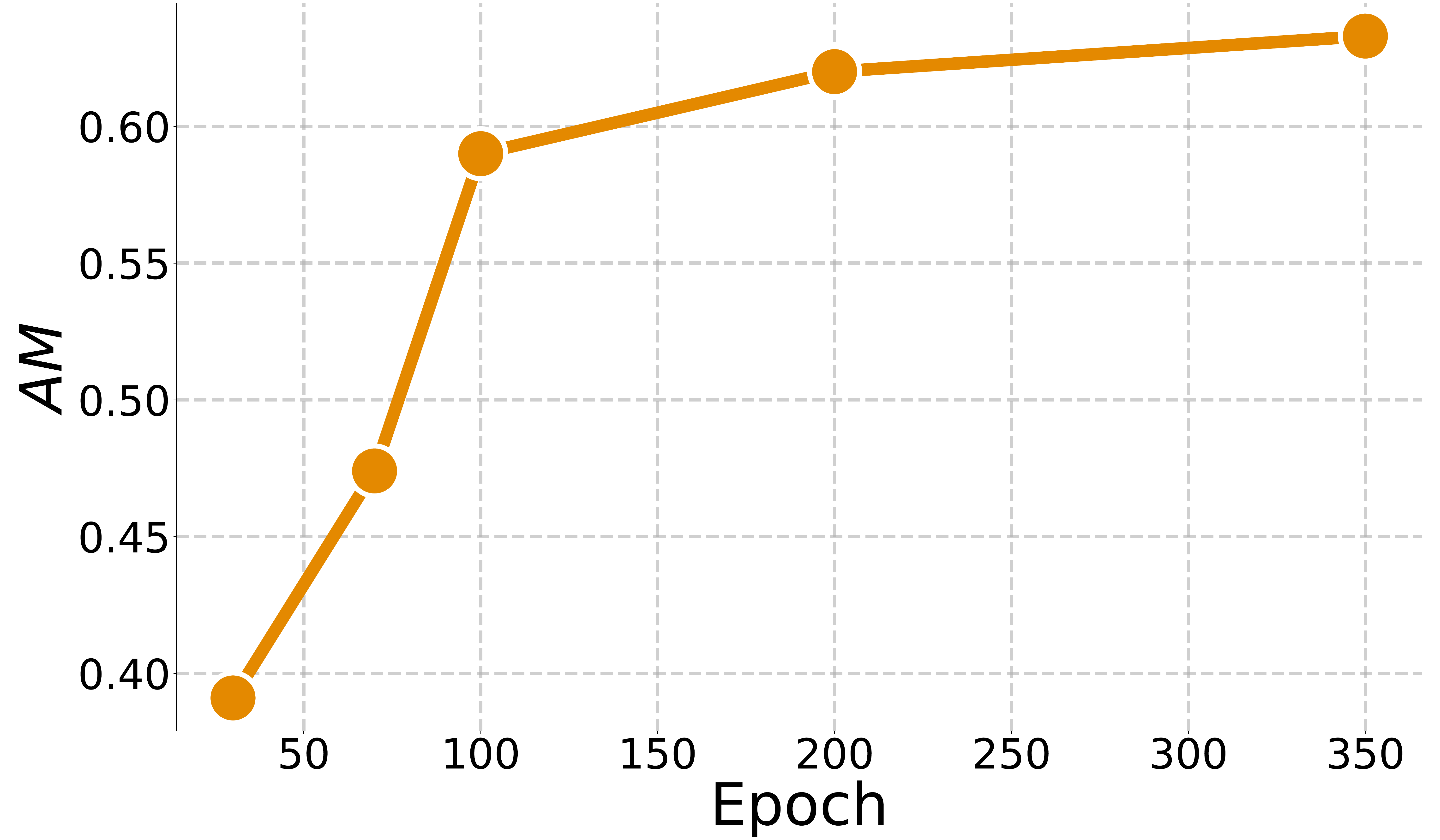}
    } 
    
    \caption{
    Compatibility of CoReS as a function of the number of training epochs  (30, 70, 100, 200, 350) at each upgrade, with 9-step upgrading. \normnummecc (a) and \avgmetr (b).
    }
    \label{fig:abl_epoch}
\end{figure}

Fig.~\ref{fig:abl_epoch} shows plots of \normnummecc (Fig.~\ref{fig:MECC_at_epoch}) and \avgmetr (Fig.~\ref{fig:M_avg_at_epoch}) for this experiment. We can notice that the trade-off between \normnummecc and \avgmetr occurs near 100 epochs. As \normnummecc decreases, the \avgmetr loses its meaning as a metric for compatibility evaluation.

\subsection{Model Selection}
\label{sec:model_selection}
In Sec.~\ref{sec:extensions_cores}, we considered the case in which at each upgrade the training-set is increased in size and extended the basic CoReS formulation introducing compatibility learning with Model Selection. Tab.~\ref{tab:hyper_tuning} compares \normnummecc and \avgmetr of the basic CoReS implementation against CoReS with Model Selection on \cifar-100 and nine upgrade steps in 20 runs. 
It can be observed that Model Selection provides some improvement of \normnummecc while keeping the same \avgmetr. 
A larger improvement is observed when Model Selection combined with pre-training as discussed in the next subsection.

\begin{table} [t]
\small
\centering 
\caption{
Compatibility of CoReS  for training from scratch without Model Selection (\textsc{CoReS w/oMS}) and with Model Selection (\textsc{CoReS wMS}) at each upgrade, with 9-step upgrading over 20 runs. Mean and standard deviation of \normnummecc and \avgmetr.
} 
\label{tab:hyper_tuning} 

\setlength{\tabcolsep}{8pt}

\begin{tabular}{lcc} 
\toprule 
 & \normnummecc & \avgmetr \\

\midrule
            \textsc{CoReS w/o MS}     & 0.51 $\pm$ 0.10             & 0.59 $\pm$ 0.08 \\
            \textsc{CoReS w MS}             & 0.53 $\pm$ 0.11      & 0.59 $\pm$ 0.09 \\

\bottomrule
\end{tabular} 
\end{table}

\subsection{Pre-training}
\label{sec:pre_train}
In this section we analyze the effect of pre-training over compatibility. We reserved 50 classes of \cifar-100 to pre-train the model and 50 classes to create 10 distinct training-sets to upgrade the model. The 50-classes pre-trained model was used for initialization. Five classes were added at each upgrade. Since the size of pre-training changes over the upgrades, we expected to observe the correlation between pre-training and compatibility.

\begin{table} [t]
    \centering
    \caption{Compatibility  of CoReS for training from a pre-trained model without (\textsc{w/o MS}) and with (\textsc{w MS}) Model Selection at each upgrade, with 9-step upgrading over 20 runs. Mean and standard deviation of \normnummecc and \avgmetr as a function of the number of classes in the pre-trained model. 
    }
    \label{tab:pretrain}
    \setlength{\tabcolsep}{4pt}
    \begin{tabular}{c  cccc } 
        \toprule 
        \multirow{3}{*}{\shortstack{\textsc{Pre-training} \\ \textsc{\#classes}}} 
        & \multicolumn{2}{c}{\textsc{w/o MS}}
        & \multicolumn{2}{c}{\textsc{w MS}} 
        \\ 
        
        \cmidrule(lr){2-3} \cmidrule(lr){4-5} 
        &\normnummecc  & \avgmetr       
        &\normnummecc  & \avgmetr  
        
        \\ \midrule
        
        20   & 0.56 $\pm$ 0.16 & 0.65 $\pm$ 0.13 & 0.64 $\pm$ 0.11 & 0.64 $\pm$ 0.09  \\ 
        30   & 0.60 $\pm$ 0.14 & 0.67 $\pm$ 0.14 & 0.68 $\pm$ 0.09 & 0.65 $\pm$ 0.12  \\ 
        40   & 0.62 $\pm$ 0.19 & 0.69 $\pm$ 0.09 & 0.71 $\pm$ 0.12 & 0.67 $\pm$ 0.08 \\ 
        50   & 0.67 $\pm$ 0.13 & 0.70 $\pm$ 0.12 & 0.74 $\pm$ 0.14 & 0.69 $\pm$ 0.11 \\ 
        \bottomrule
    \end{tabular}
\end{table}

The values of \avgmetr and \normnummecc in relationship with the number of classes leveraged for pre-training are reported in Tab.~\ref{tab:pretrain} for CoReS without and with Model Selection over 20 runs. From the table, it appears that the more data are used for pre-training the more the model improves the \normnummecc and the \avgmetr. 
The Model Selection strategy provides an additional increase of \normnummecc of about 8\% at any level of pre-training with nearly the same improvement of \avgmetr.

\begin{figure*}[t]
    \centering
    \begin{minipage}{0.76\textwidth}
        \hspace{-0.45cm}
        \subfigure[CoReS]{ \label{fig:10_step_CORES_MS}
            \includegraphics[width=0.49\linewidth]{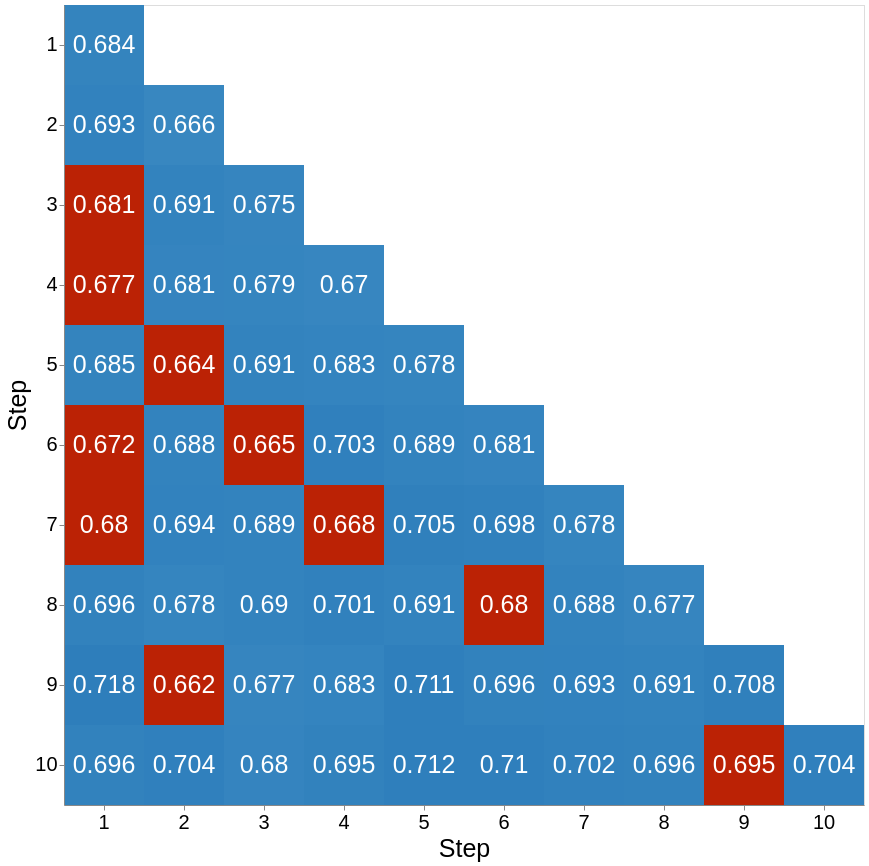}
        }
        \subfigure[BCT~\cite{DBLP:conf/cvpr/ShenXXS20}]{ \label{fig:10_step_BCT_MS}
            \includegraphics[width=0.49\linewidth]{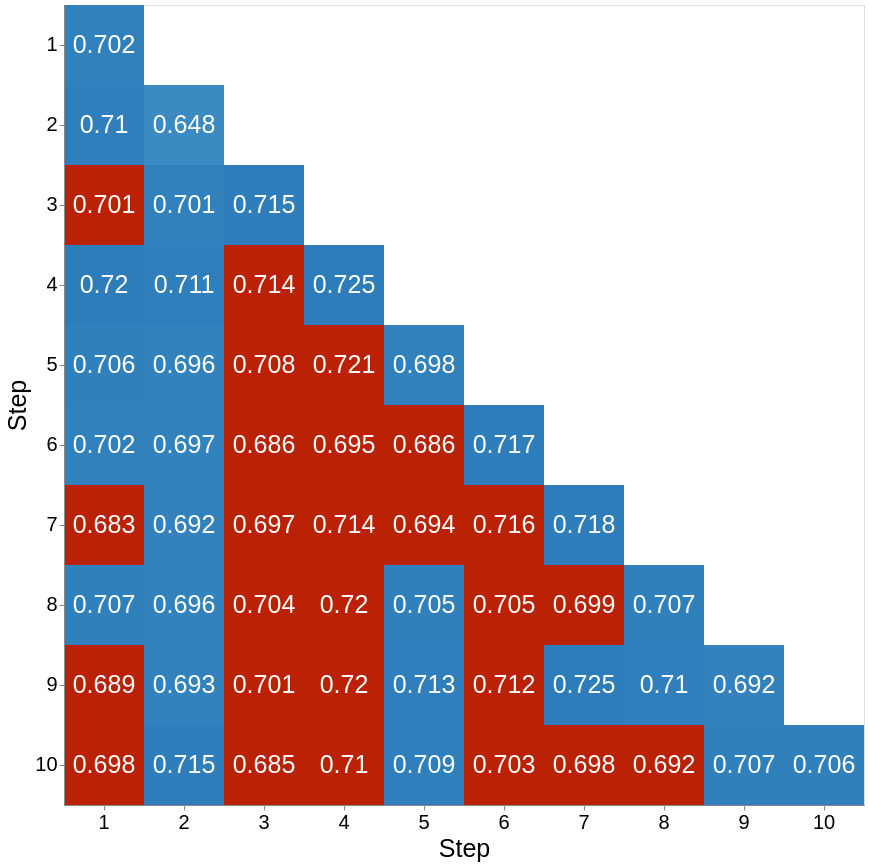}
            \includegraphics[width=0.034\linewidth]{imgs/10step/bar10.png}
        }
    \end{minipage}
    \caption{Compatibility matrices of (a) CoReS  and  (b) BCT with 9-step upgrading. Models are pre-trained with 50 \cifar-100 classes and the Model Selection strategy is used.   
    }
    \label{fig:comp_classes}
\end{figure*}

For the sake of comparison, we implemented pre-training with Model Selection also on BCT in the same test scenario. The compatibility matrices of CoReS and BCT are shown in Fig.~\ref{fig:10_step_CORES_MS} and  Fig.~\ref{fig:10_step_BCT_MS}, respectively. We see that CoReS achieves compatibility in almost all the upgraded models, achieving an \normnummecc of $0.78$, with an evident large improvement over BCT. 

\begin{table} [t]
\small
\centering 
\caption{ 
Compatibility of CoReS for different model initializations at each upgrade, with 9-step upgrading over 20 runs:
with same random initialization  (\textsc{Same}); with a different random initialization (\textsc{Random}); with fine tuning from the previously learned model (\textsc{Fine-tuned}). 
Mean and standard deviation of \normnummecc and \avgmetr.} 
\label{tab:ablation_other_settings} 

\setlength{\tabcolsep}{12pt}

\begin{tabular}{lcc} 
\toprule 

\textsc{Initialization } & \normnummecc & \avgmetr \\

\midrule
            \textsc{Same}    & 0.51 $\pm$ 0.10      & 0.59 $\pm$ 0.08 \\
            \textsc{Random}   & 0.45 $\pm$ 0.09      & 0.60 $\pm$ 0.11 \\
            \textsc{Fine-tuned}      & 0.38 $\pm$ 0.12      & 0.57 $\pm$ 0.13 \\

\bottomrule
\end{tabular} 
\end{table}

\subsection{Model Initialization} 
\label{sec:Different Model Initialization}

In our previous experiments, the network was trained from scratch at each upgrade, using the same, randomly selected, initial parameters. According to this, all the upgraded models started  optimization from the same configuration of parameters. Alternatively, parameters can be randomly selected at each upgrade. As a third option, learning at each upgrade can also be performed by starting from the previous upgrade and then performing fine-tuning incrementally.

Tab.~\ref{tab:ablation_other_settings} reports mean and standard deviation of \normnummecc and \avgmetr for 9-step upgrading and 20 runs for these three cases (\textsc{Same}, \textsc{Random}, and \textsc{Fine-tuned}, respectively). It can be noticed that random initialization at each upgrade negatively affects the final performance, resulting into a sensible reduction of \normnummecc with respect to initialization with the same parameters at each upgrade. This result can be supported by the observations in \cite{NEURIPS2019_05e97c20,neyshabur2020being} and is likely related to the concept of ``flat minima'' \cite{hochreiter1995simplifying,hochreiter1997flat} and to the fact that with the same initialization, the SGD optimization explores the same basin of the loss landscape near the minimum. 
We argue that the basins of the loss landscape near the minimum explored by CoReS across upgrades has close relationship with the stationarity/compatibility of the learned representation. 
On its turn, \textsc{Fine-tuned} results into a strong reduction in $AC$. We argue that this is due to the fact that in this case, differently from learning from scratch, optimization should go back to earlier weight configurations to find more complex error landscapes.

\begin{table} [t]
\small
\centering 
\caption{Compatibility of CoReS as a function of sequential class ordering with 9-step upgrading over 20 runs:
with alphabetical order (\textsc{Alphabetical}); with different random permutation of the classes at each run (\textsc{Random}). Mean and standard deviation of \normnummecc and \avgmetr.
} 
\label{tab:ablation_other_order} 

\setlength{\tabcolsep}{8pt}

\begin{tabular}{lcc} 
\toprule 
\textsc{Class Ordering} & \normnummecc & \avgmetr \\

\midrule
            \textsc{Alphabetical}    & 0.51 $\pm$ 0.10      & 0.59 $\pm$ 0.08 \\
            \textsc{Random}              & 0.52 $\pm$ 0.13      & 0.57 $\pm$ 0.05 \\

\bottomrule
\end{tabular} 
\end{table}

\label{sec:cores_IFT}

\subsection{Different class order} 
\label{sec:different_order_classes}
Tab.~\ref{tab:ablation_other_order} reports mean and standard deviation of \normnummecc and \avgmetr of 20 runs for different sequential class orderings, namely: (1) classes are alphabetically ordered, (\textsc{Alphabetical}); (2) each run has a different random permutation of the classes (\textsc{Random}), which we adopted as default choice in our experiments. It can be noticed, \normnummecc and \avgmetr metrics are almost the same in the two cases showing that class ordering has essentially no effect on learning compatible features.

\subsection{Different Model Architectures}
\label{sec:Different Model Architecture}
It has been shown that the expressive power of the network architecture has impact on the classification accuracy of fixed classifiers since the complexity of learning compatible representations is fully demanded to the internal layers of the neural network \cite{DBLP:conf/iclr/HofferHS18,perniciTNNLS2021}. 

According to this, it is relevant to evaluate the impact of the expressive power of network architectures over CoReS compatible learning. We compared  different architectures with increasing expressive power, namely: ResNet20 (0.27M parameters), ResNet32 (0.46M parameters), SENet-18 (1.23M parameters) and  RegNetX\_200MF (2.36M parameters)\footnote{Code source for ResNet20 and RegNetX\_200MF can be found at \url{https://github.com/kuangliu/pytorch-cifar} and for ResNet32 at \url{https://github.com/arthurdouillard/incremental\_learning.pytorch}}.
We set the SGD initial learning rate to 0.1 and the weight decay factor to $5 \cdot 10^{-4}$.
ResNet20 was trained for 40 epochs with constant learning rate. ResNet32 was trained for 70 epochs, with learning rate reduced to 0.01 at epoch 50, and to 0.001 at epoch 65. RegNetX\_200MF was trained for 150 epochs, with learning rate reduced to 0.01 at epoch 70 and to 0.001 at epoch 120.

\begin{figure}[t]
    \centering
    
    \subfigure[ResNet20] {\label{fig:resnet20}
        \includegraphics[width=0.35\linewidth]{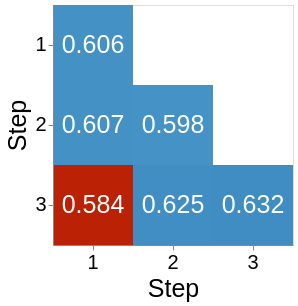}
    }
    \hspace{10pt}
    \subfigure[ResNet32] {\label{fig:resnet32}
        \includegraphics[width=0.35\linewidth]{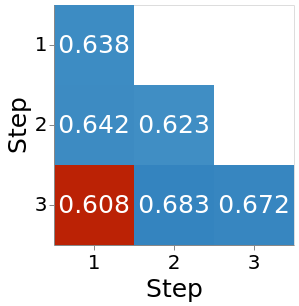}
    }
    \subfigure{
        \subfigcapmargin = 0pt
        \includegraphics[width=0.077\linewidth]{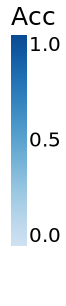}
        \addtocounter{subfigure}{-1}
    }
    \\
    \centering
    \subcapcentertrue
    \subfigure[SENet-18] {\label{fig:senet}
        \includegraphics[width=0.35\linewidth]{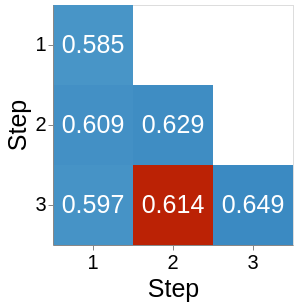}
    }
    \hspace{10pt}
    \subfigure[RegNetX\_200MF] {\label{fig:regnet}
        \includegraphics[width=0.35\linewidth]{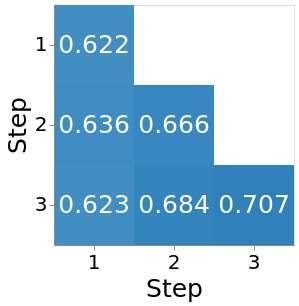}
    }
    \subfigure{
        \subfigcapmargin = 0pt
        \includegraphics[width=0.077\linewidth]{imgs/3step/bar3_01range.png}
    }
    \caption{
    Compatibility matrices of CoReS for different network architectures with 2-step upgrading:  (a) ResNet20; (b) ResNet32; (c) SENet-18; (d) RegNetX\_200MF. 
    }
    \label{fig:cores_diff_net}
\end{figure}

Fig.~\ref{fig:resnet20}, Fig.~\ref{fig:resnet32}, Fig.~\ref{fig:senet} and Fig.~\ref{fig:regnet} show the CoReS compatibility matrices with ResNet20, ResNet32, SENet-18, and RegNetX\_200MF respectively.  It can be noticed that both compatibility and verification accuracy follow the expressive power of the network model. ResNet32 in Fig.~\ref{fig:resnet32} has compatibility similar to Fig.~\ref{fig:resnet20}, but the verification accuracy is higher. RegNetX\_200MF improves both compatibility and verification accuracy Fig.~\ref{fig:regnet}. This experiment also confirms the general applicability of our compatibility learning approach.

\begin{figure}[t]
    \centering
    \includegraphics[width=0.4\linewidth]{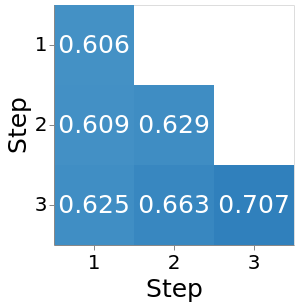}
    \includegraphics[width=0.09\linewidth]{imgs/3step/bar3_01range.png}
    \caption{
    Compatibility matrix of CoReS with a 2-step upgrading, with different network architectures: starting with ResNet20; first upgrade with SENet-18; second upgrade with RegNetX\_200MF.
    }
    \label{fig:cores_diff_net_per_step}
\end{figure}
Finally, we analyzed the practical case in which a deployed system is upgraded not only with fresh data but also with recent and more powerful network architectures. To this end, we evaluate the case in which the ResNet20 is first upgraded with SENet-18, and subsequently with RegNetX\_200MF. 
Fig.~\ref{fig:cores_diff_net_per_step}  reports the compatibility matrix for this experiment with the \cifar-100/10 datasets. As can be observed, the learned representations remain compatible as the network architecture is upgraded. The increasing expressive power of the architectures also positively reflects on compatibility.

\subsection{Different Number of Future Classes}
\label{sec:diff-numb-future-classes}
\begin{table}[t]
\caption{Ablation study on CIFAR100 for different number of future classes with 1, 2, 4, 9-step upgrading.}\label{tab:ablation_prealloc_classes}
\centering
\setlength{\tabcolsep}{2pt}
\begin{tabular}{c
                c
                c
                c
                c
                c
                c
                c
                c
}
\toprule
\multirow{3}{*}{\shortstack{\textsc{\#future} \\ \textsc{classes}}}                                 & \multicolumn{2}{c}{One-upgrade} & \multicolumn{2}{c}{Two-upgrades} & \multicolumn{2}{c}{Four-upgrades} & \multicolumn{2}{c}{Nine-upgrades} \\
\cmidrule(lr){2-3} \cmidrule(lr){4-5} \cmidrule(lr){6-7} \cmidrule(lr){8-9}
                              & \normnummecc    &  $AM$           & \normnummecc  &  $AM$         & \normnummecc   & $AM$        & \normnummecc    &  $AM$         \\
\midrule
100                                                       & 1     &  0.62       & 0.67   &  0.62     & 0.9    &  0.59    & 0.51    & 0.59     \\
200                                                       & 1    &  0.68        & 1    &  0.66        & 0.8    &  0.62      & 0.53   &  0.57      \\
500                                                       & 1     &   0.67      &   1    &   0.64       &   0.7     &  0.64    &    0.51        &   0.60     \\
1000                                                      & 1    &    0.66      & 0.67    &  0.63     & 0.9    &  0.62      &     0.56      &   0.55    \\
10000                                                     & 1    &    0.63      &   1    &   0.62     &         0.7      &  0.59   &  0.51  & 0.56    \\
\bottomrule
\end{tabular}
\end{table}

It is relevant to study how the number of future classes of Eq.~\ref{softmax_loss_virtual} influences the compatibility of a learned representation model. 
In Tab.~\ref{tab:ablation_prealloc_classes}, we evaluate CoReS with a different number of future classes with one, two, four and nine-step upgrades.
As can be noticed from the table, performance does not vary as the number of future classes increases. The $AC$ stabilizes as the number of upgrades increases. The $AC$ may have sudden variations when the number of upgrades is small as it basically averages the number of times compatibility is achieved.

A related ablation for the classification task in Class-incremental Learning has already been studied in \cite{pernici2020icpr}. CiL addresses incremental classification under catastrophic forgetting. 
Given the substantial difference between the classification and the open-set verification that is investigated in this work, the evaluation of Tab.~\ref{tab:ablation_prealloc_classes} provides stronger evidence that pre-allocating a large number of future classes in the final classification layer does not reduce or impair the general applicability of the $d$-Simplex fixed classifier \cite{perniciTNNLS2021} used to learn compatible features. 

\subsection{Different Source of Upgrade Data}
\label{sec:diff-source-upgrade-data}
  
\begin{table}[t]
    \centering
    \hspace{-10pt}
    \caption{Compatibility of CoReS for training with different source of upgrade data with 1-step upgrading: from new classes (\textsc{New classes}); from the old classes (\textsc{Old classes}); from both new and old classes (\textsc{Old \& new classes}). }
    \label{tab:different_sample_per_step}
    \setlength{\tabcolsep}{2pt}
    \begin{tabular}{c
                    c
                    c
                    c
                    c
                    c
                    c
                    c
                    c
                    c
                    } 
                    
        \arrayrulecolor{black}
        \toprule 
        \multirow{5}{*}{\shortstack{\textsc{Comparison} \\ \textsc{Pair}}} &
        \multicolumn{9}{c}{\textsc{Source of upgrade data} }
        \\
        \cmidrule(lr){2-10}
        
        & \multicolumn{3}{c}{\multirow{2}{*}{\shortstack{\textsc{New classes}}} }
        &  \multicolumn{3}{c}{\multirow{2}{*}{\shortstack{\textsc{Old classes }}}} 
        & \multicolumn{3}{c}{\multirow{2}{*}{\shortstack{\textsc{Old \& new } \\ \textsc{classes}} }}
        \\ 
        \\
         
        \cmidrule(lr){2-4}
        \cmidrule(lr){5-7} 
        \cmidrule(lr){8-10} 
        & $M$ & \textsc{ECC} & $\Gamma$ (\%) 
        & $M$ & \textsc{ECC} & $\Gamma$ (\%) 
        & $M$ & \textsc{ECC} & $\Gamma$ (\%) 
        \\
        
        \midrule
        $(\phi_{\rm old}, \phi_{\rm old})$  & 0.60 & \textendash  & \textendash  & 0.64 & \textendash  & \textendash  & 0.59 & \textendash  & \textendash   \\
        $(\phi_{\rm new}, \phi_{\rm old})$  & 0.61 & $\surd$ & 21.3   & 0.65 & $\surd$ & 33.3 & 0.64 & $\surd$ & 55.6 \\
        $(\phi_{\rm new}, \phi_{\rm new})$  & 0.65 & \textendash  & \textendash   &  0.66 & \textendash  & \textendash &   0.66 & \textendash  & \textendash   \\ 
        \bottomrule
    \end{tabular} 
\end{table}
In this section we study how performance changes if new data used to upgrade the models belongs to some of the already seen classes.
In Tab.~\ref{tab:different_sample_per_step} ``\textsc{Old classes}'' represents this new case in which $\phi_{\rm old}$ is trained on the 100\% of classes with the 50\% of samples per class and upgraded using the 100\% of training classes with 100\% of samples per class; the ``\textsc{New classes}'' column refers to the standard case reported in Sec. 4.1 of the submitted manuscript in which the old model $\phi_{\rm old}$ is first trained using 50\% of classes with 100\% of samples per class, while the upgraded model $\phi_{\rm new}$ using the 100\%  of classes with 100\% of samples; and finally ``\textsc{Old \& new classes}'' indicates the performance of training $\phi_{\rm old}$ using 50\% classes with 50\% of samples per class while the upgraded model $\phi_{\rm new}$ is trained using the 100\% of classes with 100\% of samples. 
As can be seen from the table, CoReS achieves a compatible representation between $\phi_{\rm old}$ and $\phi_{\rm new}$ in all of the three scenarios, showing that it is effective also in the case in which the upgrade data come from already seen classes. This is substantially due to the fact that the $d$-Simplex fixed classifier introduced in \cite{perniciTNNLS2021} is less susceptible to the reduced (i.e., imbalanced) number of data samples as recently shown in \cite{yang2022we} and in \cite{kasarla2022maximum}. 

\section{Conclusions}
In this paper, we presented CoReS, a new training procedure that provides compatible representations grounding on the stationarity of the representation as provided by fixed classifiers. In particular, we exploited a special class of fixed classifiers where the classifier weights are fixed to values taken from the coordinate vertices of regular polytopes. They define classes that are maximally separated in the representation space and maintain their spatial configuration stationary as new classes are added. 
With our solution, there is no need to learn any mappings between representations nor to impose pairwise training with the previously learned model. The functional complexity is fully demanded to the internal layers of the network. We analyzed and discussed the factors that affect the performance of our method and performed extensive comparative experiments on visual search applications of different complexity. We demonstrated that our solution for compatibility learning based on feature stationarity largely outperforms the current state-of-the-art and is particularly effective in the case of multiple upgrades of the training-set, that is the typical case in real applications. CoReS also provides large compatibility when the network architecture is upgraded to newer and more powerful models.

\ifCLASSOPTIONcompsoc
  \section*{Acknowledgments}
\else
  \section*{Acknowledgment}
\fi

This work was partially supported by the European Commission under European Horizon 2020 Programme, grant number 951911 - AI4Media.

The authors also acknowledge the CINECA award under the ISCRA initiative (ISCRA-C - ``ILCoRe'', ID:~HP10CRMI87), for the availability of high-performance computing resources and thank Giuseppe Fiameni (Nvidia) for his support.

\ifCLASSOPTIONcaptionsoff
  \newpage
\fi

\bibliographystyle{unsrt}
\bibliography{bibliography}

\vspace{-1cm}
\begin{IEEEbiography}[{\includegraphics[width=1in,height=1.25in,clip,keepaspectratio]{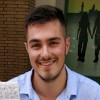}}]{Niccolò Biondi}
Niccolò Biondi received the M.S. degree (cum laude) in Computer Engineering from the University of Firenze, Italy, in 2021. Presently, he is a PhD. student at the University of Firenze at MICC, Media Integration and Communication Center, University of Firenze. His research interests include machine learning and computer vision with special focus on compatible learning, representation learning, and incremental learning.
\end{IEEEbiography}
\vskip 0pt plus -1fil
\vspace{-1cm}
\begin{IEEEbiography}
[{\includegraphics[width=1in,height=1.25in,clip,keepaspectratio]{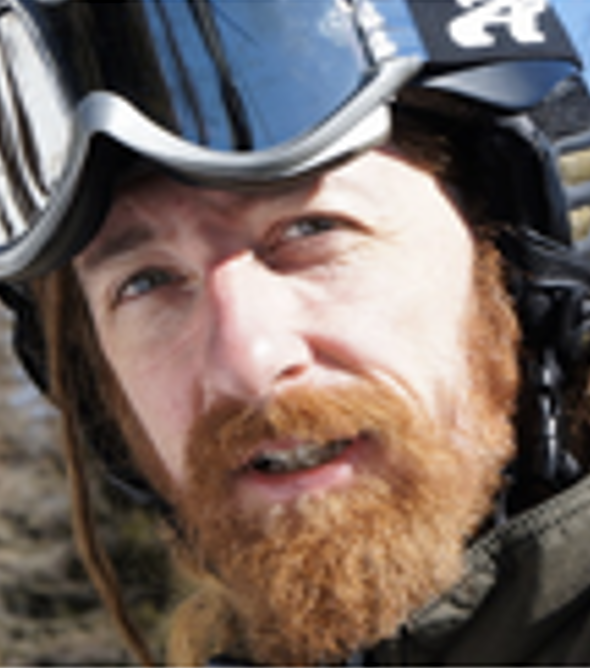}}]{Federico Pernici}
Federico Pernici is Assistant Professor at the University of Firenze, Italy. He received the laurea degree in Information Engineering in 2002, the post-laurea degree in Internet Engineering in 2003 and the Ph.D. in Information and Telecommunication Engineering in 2005 from the University of Firenze, Italy. He has been working at the MICC Media Integration and Communication Center as a research assistant since 2002, and has also served as adjunct professor at the University of Firenze. His scientific interests are computer vision and machine learning with a focus on different aspects of visual tracking, incremental learning and representation learning. He was Associate Editor of Machine Vision and Applications journal. 
\end{IEEEbiography}
\vskip 0pt plus -1fil
\vspace{-1cm}
\begin{IEEEbiography}
  [{\includegraphics[width=1in,height=1.25in,keepaspectratio]{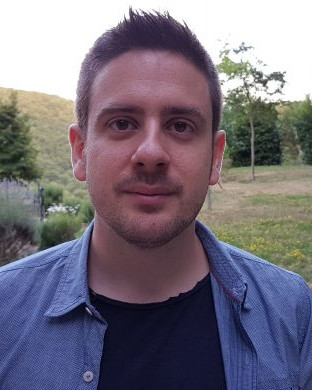}}]{Matteo Bruni}
Matteo Bruni received the M.S. degree (cum laude) in Computer Engineering 
from the University of Firenze, Italy, in 2016, the Ph.D. in Information Engineering at the University of Firenze in 2020. His research interests include pattern recognition and computer vision with specific focus on feature embedding, face recognition, incremental learning and compatible representation.
\end{IEEEbiography}
\vskip 0pt plus -1fil
\vspace{-1cm}
\begin{IEEEbiography}
  [{\includegraphics[width=1in,height=1.25in,clip,keepaspectratio]{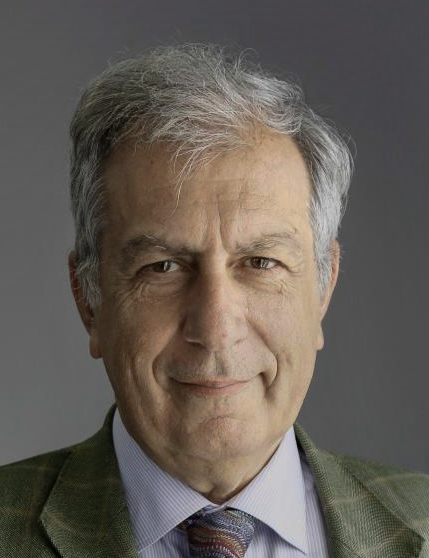}}]{Alberto Del Bimbo}
Prof. Del Bimbo is Full Professor at the University of Firenze, Italy. He is the author of over 350 scientific publications in computer vision and multimedia and principal investigator of technology transfer projects with industry and governments. He was the Program Chair of ICPR 2012, ICPR 2016 and ACM Multimedia 2008, and the General Chair of IEEE ICMCS 1999, ACM Multimedia 2010, ICMR 2011, ECCV 2012 and ICPR2020. He is the General Chair of ACM Multimedia 2021. He is the Editor in Chief of ACM TOMM Transactions on Multimedia Computing Communications and Applications and Associate Editor of Multimedia Tools and Applications and Pattern Analysis and Applications journals. He was Associate Editor of IEEE Transactions on Pattern Analysis and Machine Intelligence, IEEE Transactions on Multimedia and Pattern Recognition and also served as the Guest Editor of many Special Issues in highly ranked journals. Prof. Del Bimbo is the Chair of ACM SIGMM the Special Interest Group on Multimedia. He is IAPR Fellow and ACM Distinguished Scientist and is the recipient of the 2016 ACM SIGMM Award for \emph{Outstanding Technical Contributions to Multimedia Computing Communications and Applications}.
\end{IEEEbiography}

\end{document}